\documentclass{article}

\usepackage[preprint]{neurips_2026}

\usepackage[utf8]{inputenc} 
\usepackage[T1]{fontenc}    
\usepackage[hidelinks]{hyperref}       
\usepackage{url}            
\usepackage{booktabs}       
\usepackage{amsfonts}       
\usepackage{nicefrac}       
\usepackage{microtype}      
\usepackage{xcolor}         

\usepackage[ruled, vlined]{algorithm2e}

\usepackage{amsmath}
\usepackage{amssymb}
\usepackage{amsthm}
\usepackage{optidef}
\usepackage{tcolorbox}
\usepackage{xfrac}
\usepackage{placeins}
\usepackage{float}

\usepackage{fontawesome5}

\usepackage{bbm}

\newtheorem{theorem}{Theorem}
\newtheorem{lemma}[theorem]{Lemma}

\newtheorem{proposition}[theorem]{Proposition}
\newtheorem{definition}[theorem]{Definition}

\DeclarePairedDelimiter{\norm}{\lVert}{\rVert}
\DeclarePairedDelimiter{\br}{\lparen}{\rparen} 
\DeclarePairedDelimiter{\sq}{\lbrack}{\rbrack} 
\DeclarePairedDelimiter{\cu}{\lbrace}{\rbrace} 

\renewcommand{\*}{\mathcal}

\newcommand{\E}{\mathbb E}

\DeclareMathOperator{\Span}{span}

\newcommand{\textcmrbold}[1]{{\fontfamily{cmss}\bfseries\selectfont #1}}
\newcommand{\textcmr}[1]{{\fontfamily{cmss}\selectfont #1}}

\title{Probing for Representation Manifolds in Superposition}

\author{%
  Alexander Modell \\
  Department of Mathematics\\
  Imperial College London, U.K.\\
  \texttt{a.modell@imperial.ac.uk} \\
} 

\begin{document}

  \maketitle

\begin{abstract}
  This paper introduces the Manifold Probe, a supervised method for discovering representation manifolds in superposition. The method generalizes linear regression probes by learning the space of features of a concept that can be linearly predicted from the representations, and then learning the directions used to encode them. We demonstrate the probe on representations of time and space in Llama 2-7b, finding manifolds which linearly represent an interpretable set of features in each case. In the case of time, we show that by steering along the manifold, we can influence the model's completions about the years in which famous songs, movies and books were released, providing evidence that the Manifold Probe can discover manifolds which are causally involved in model behaviour. 

  \begin{center}
    \href{https://github.com/alexandermodell/maniprobe}{\tt \faGithub\ alexandermodell/maniprobe}
    \vspace{-1em} 
  \end{center}
\end{abstract}

\section{Introduction}

The ability to interpret the representation geometry of large language models is a fundamental goal in a larger scientific effort to understand AI systems as a whole \citep{bereska2024mechanistic}.

A key hypothesis in this effort is the linear representation hypothesis \citep{nanda_emergent_2023}: that neural networks organize their internal representations so as to make semantically important features accessible via linear projections. A related hypothesis is that of superposition \citep{smolensky_tensor_1990, mikolov_linguistic_2013, elhage_toy_2022}: the idea that representations of distinct concepts combine additively to produce representations of their joint semantics. 

Initial work around these hypotheses focused on understanding representations of simple binary concepts, which are considered to be either present or absent \citep{elhage_toy_2022}. In this setting, concepts are hypothesised to be represented by almost-orthogonal directions in representation space and the presence of multiple features is represented by summing the corresponding directions. These hypotheses motivate many contemporary interpretability methodologies, such as linear probes \citep{alain_understanding_2017,nanda_emergent_2023}, sparse autoencoders \citep{elhage_toy_2022, bricken_towards_2023, cunningham2024sparse}, and steering vectors \citep{li2023inference, marks2024geometry, rimsky2024steering, turner2025steering}.

More recently, there has been a push towards understanding the representation geometry of more complex, continuous concepts, which don't fit in to the binary framework. Examples include numerics, time, space, colour, and more abstract concepts such as emotion, ideaology and phylogeny \citep{gurnee2024language, olah_what_2024, engels_not_2025, modell2025origins, gurnee2025when, pearce2025tree, savietto2026geometry, choi2026latent, sofroniew2026twheemotion, sun2026valence}.
There is growing empirical evidence that continuous concepts are represented on manifolds which bend and twist through multiple dimensions of the representation space \citep{gorton_curve_2024, modell2025origins, yocum2025neural, gurnee2025when, karkada2026symmetry}. The presence of such multi-dimensional manifolds is compatible with both the linear representation and superposition hypotheses. In particular, their shape directly determines the information about the concept which can be accessed via linear projections.

The problem of isolating multidimensional representations of concepts, and discovering the geometry of representation manifolds in superposition is relatively unexplored. For the former, we are only aware of \citet{engels_not_2025} who propose to do this by clustering dictionary vectors in sparse autoencoders. For the latter, \citet{yocum2025neural} and \citet{gurnee2025when} propose fitting a family of linear classifier probes to a discretization of the concept space, and \citet{modell2025origins} propose approximating representation manifolds with neighbour graphs. 

In this paper, we propose the Manifold Probe, a supervised probing methodology to discover representation manifolds which are represented in superposition with other semantic information. 

The probe is fitted in two stages. The first stage is to learn the space of features $f(z)$ of the concept values $z$ which are well-predicted by a linear function $w^\top x + b$ of the representations $x$. While a standard linear regression probe would consider a fixed feature as its target, our probe learns the features at the same time as the regression parameters. 
We show that under a generic statistical model of a representation manifold in superposition, these learned features approximate the geometry of the manifold with respect to some unknown basis. The second stage learns this basis using linear regression.
The main methodological contribution of this paper is the formulation and optimization of the first step of this procedure. 

To demonstrate the Manifold Probe, we use it to explore residual-stream representations of time and space in Llama 2-7b from probing datasets curated in \citet{gurnee2024language}. While \citet{gurnee2024language} show that the concept values themselves are linearly represented, our probe brings to light many more features which are too, some of which are represented more precisely than the concept values themselves.

We show how applying factor analysis to the learned features can help us interpret them. Employing a Varimax rotation which aims to make the features approximately sparse reveals that the time manifold linearly separates decades, while the space manifold linearly separates many states in the U.S.A..

Finally, we show that the time manifold we find is not only present in the residual stream, but is used by the model.
We perform an intervention experiment where, at a given layer, we steer the residual stream representations by adding steering vectors which trace the manifold. By doing this, we can influence the model to complete a prompt about the release date of a song, movie or book with a year that we target.

\section{Setup and background}

\subsection{Concepts and representation manifolds}

A \emph{concept} is a topological space $\*Z$ which we can attach some real world meaning to. The simplest example might be a binary concept which is considered to be either present or absent. 
Continuous concepts include time (a line $\*Z = \mathbb R$), space (a plane $\*Z = \mathbb R^2$ or a sphere $\*Z = \mathbb S^2$), colour (a cylinder $\*Z = \mathbb S \times \mathbb R^2$ or cube $\*Z = \mathbb R^3$) and can include more abstract concepts such as emotion with an appropriate mathematical model (such as the valence-arousal-dominance model). 

We say that any injective map $\phi : \mathcal Z \to \mathbb R^p$ \emph{represents} the concept $\*Z$.
If $\phi$ is also continuous, then its image $\*M = \phi(\*Z)$ is a \emph{representation manifold} embedded in $\mathbb R^p$ which, under some mild conditions\footnote{such as $\*Z$ being compact.}, is topologically equivalent to the concept $\*Z$. We'll write $\*U \subseteq \mathbb R^p$ to denote the smallest subspace containing $\*M$, and $d$ to denote its dimension.

For example, if $\*Z$ is an interval, then $\*M$ is a curve; if $\*Z$ is a circle, then $\*M$ is a loop; and if $\*Z$ is a rectangle, then $\*M$ is a sheet, all of which might bend and twist to occupy more dimensions in the representation space than might be implied by the intrinsic dimensionality of the concept itself.

\subsection{Semantics and superposition}

We now turn to the question of how multiple concepts might be represented together.

To this end, we will consider an abstract topological space $\*S$ which we refer to as the \emph{semantic space}, which we assume encodes the semantics of any input to the neural network. We will assume $\*S$ can be factorized into a set of interpretable concepts $\*Z_1, \cdots, \*Z_m$, and a set of other semantics $\Xi$, so that
\[
\*S = \*Z_1 \times \cdots \times \*Z_m \times \Xi.
\]

We will be interested in hypothesizing about, and making inferences about the structural form of a map $x : \*S \to \mathbb R^p$ which represents $\*S$.

A key hypothesis in mechanistic interpretability is that of \emph{superposition}: the idea that representations of concepts combine additively to produce representations of their joint semantics. 

\begin{definition}
  \label{def:superposition}
  We say that a map $x: \*S \to \mathbb R^p$ represents the concepts $\*Z_1, \cdots, \*Z_m$ in superposition if there exists maps $\phi_i : \*Z_i \to \*U_i \subseteq \mathbb R^p$ for $i = 1, \cdots, m$, and a map $\eta : \Xi \to \*V \subseteq \mathbb R^p$such that
  \begin{equation}
    \label{eq:superposition}
    x(s) = \phi_1(z_1) + \cdots + \phi_m(z_m) + \eta(\xi)
  \end{equation}
  for all $s = (z_1,\ldots,z_m, \xi) \in \*S$.
\end{definition}

In the special case that concept representations are one-dimensional, this superposition hypothesis has been studied extensively in the mechanistic interpretability literature.
The general setting presented above, in which concept representations are allowed to occupy multidimensional subspaces, has received comparatively little attention, with the notable exception of \citet{engels_not_2025}. It also presents an additional inference problem: not only is it of interest to estimate the subspace which the concept representation occupies, but also the geometry of the representation \emph{within} that subspace.

In this paper, we will be concerned with developing methodology to discover the representation $\phi := \phi_1$ of a single target concept $\*Z := \*Z_1$. From hereon, we will absorb any additional concepts in $\Xi$, and assume that $\*S = \*Z \times \Xi$. 

\subsection{Probing}

In order to discover the representation $\phi$ of the target concept $\*Z$, we will employ the probing methodology \citep{alain2016understanding}. The idea behind probing is to construct a dataset of representation-concept values pairs $\*D = \{(x_i, z_i)\}_{i=1}^n$, and to use this in a supervised fashion to fit a statistical model which elucidates the representation geometry of interest.

We assume that each representation-concept value pair $(x_i,z_i) \sim \mathsf P$ in the probing dataset $\*D$ is sampled independently by sampling a semantic value $s_i = (z_i, \xi_i)$ from a distribution $\mathsf P_{\*S}$, and then constructing the representation $x_i = x(s_i)$ according to the superposition equation \eqref{eq:superposition} in Definition~\ref{def:superposition}. While we observe the representation-concept value pairs $(x_i, z_i)$; the nuisance semantics $\xi_i$, and the functional form of the maps $\phi$ and $\eta$ are unobserved.
We assume that the concept value $z_i$ and nuisance semantics $\xi_i$ are independent, i.e. $\mathsf P_{\*S} = \mathsf P_{\*Z} \times \mathsf P_{\Xi}$.

Our statistical objective is to use the probing dataset $\*D$ to learn two maps which estimate $\phi(z)$ from either the concept value $z$, or a corresponding representation $x$:
\begin{enumerate}
  \item a smooth non-linear map $\hat \phi : \*Z \to \hat{\*M} \subset \mathbb R^p$ from the concept values $\*Z$ to a manifold $\hat{\*M}$ in some $d$-dimensional subspace $\hat{\*U} \subset \mathbb R^p$ of the representation space.
  \item a linear (affine) map $\Psi : \mathbb R^p \to \hat{\* U} \subset \mathbb R^p$ from the representation space $\mathbb R^p$ to the subspace $\hat{\*U} \subset \mathbb R^p$.
\end{enumerate}

We point out at this stage that the maps $\phi$ and $\eta$ in the superposition equation \eqref{eq:superposition} are only defined up to translation, and so for the purpose of estimation, we will assume without loss of generality that $\mathbb E[\phi(z)] = 0$.

\subsection{Manifold estimation as regression}

Before discussing how we might estimate the representation map $\phi$ from a finite probing dataset $\*D$, it is useful to consider how we might obtain $\phi$ given access to the true underlying population distribution $\mathsf P$.

The following lemma, which we prove in Section~\ref{sec:proof_of_population_regression} of the appendix, tells us how.

\begin{lemma}
  \label{lem:population_regression}
  There exists a basis $u_1,\ldots,u_d \in \*U$ and a set of mean-zero, orthonormal features $f_1,\ldots,f_d : \*Z \to \mathbb R$ such that
  \begin{equation}
    \label{eq:decomposition}
    \phi(z) = u_1 f_1(z) + \ldots u_d f_d(z)
  \end{equation}
  which also solve the following sequential population regression problems:
    \begin{align}
      (f_k, w_k, b_k) &= \underset{\substack{f : \*Z \to \mathbb R, \; w \in \mathbb R^p, \; b\in \mathbb R \\ \mathbb E(f) = 0, \mathbb E(f^2) = 1 \\ f \perp f_{k-1}, \ldots, f_1}}{\operatorname{argmin}} \mathbb E\left[ (f(z) - w^\top x - b)^2\right],
      \label{eq:f_objective}
      \\
      (u_k, c_k) &= \qquad \underset{u, c \in \mathbb R^p}{\operatorname{argmin}}\qquad  \mathbb E\left[\left\|x - u f_k(z)- c\right\|^2\right].
      \label{eq:u_objective}
    \end{align}
  where expectations are taken with respect to $(x,z) \sim \mathsf P$, and $f \perp g$ means $\mathbb E(fg) = 0$.
\end{lemma}

Lemma~\ref{lem:population_regression} suggests the shape of a representation manifold is intimately connected to space of features which it linearly represents. This dual interpretation is key to our finite-sample estimation procedure, and also provides a lens through which to interpret the manifold geometry.

\section{Methodology}

This section is dedicated to developing a sample-based estimation procedure for estimating $\phi$ from the probing data $\*D$ based on the population regression problems in Lemma~\ref{lem:population_regression}.

In order to fit a feature $f$, we parametrize it in some basis $h_1,\ldots,h_m$ which we treat as known, so that it can be written as
\begin{equation}
  \label{eq:f_parametrization}
  f(z) = \beta^\top h(z) \equiv \beta_1 h_1(z) + \cdots + \beta_m h_m(z)
\end{equation}
for some unknown scalar parameters $\beta := (\beta_1,\ldots,\beta_m)$. We denote the space of functions of the form \eqref{eq:f_parametrization} as $\mathcal{H}$. An appropriate choice of basis depends on the nature of the concept space $\*Z$. In our examples, we use cubic B-splines, or tensor products thereof, however our method can accomodate any choice of basis.

In order to avoid overfitting the function $f$, we need some way to control its complexity. The standard approach in functional regression is to choose an overly flexible basis, and then to control the complexity of $f$ via a penalty function $J(f)$ which we add to our loss function. The advantage of this approach is that it allows us to choose the level of permitted complexity using the data.

In this paper, we will assume that the chosen penalty function $J$ is quadratic which allows us to write it as a quadratic form $J(f) = \beta^\top S \beta$ in the basis coefficients $\beta$. In our examples, we use the integrated, squared second derivative penalty
\[
  J(f) = \int_{\*Z} [f''(z)]^2 \;\mathsf d z,
\]
which is usually considered a default choice. However, our method is flexible enough to accomodate quadratic penalty, and in Section~\ref{sec:selecting_regularization_parameters} of the appendix, we discuss how our method can be adapted to accomodate non-quadratic penalties, such as $\ell_1$ and mixed $\ell_1$ and $\ell_2$-type penalties.

With $\*H$ and $J$ defined, we are ready to write down our probing procedure for estimating $\phi(z)$. 

\begin{definition}
  \label{def:manifold_probe}
  We write $\hat \phi, \Psi = \textsf{ManifoldProbe}(\*D, d; \lambda_w, \lambda_f)$ if
  \[
    \hat \phi(z) = \hat u_1 \hat f_1(z) + \cdots + \hat u_d \hat f_d(z), \qquad \Psi(x) = \hat u_1 g_1(x) + \cdots + \hat u_d g_d(x),
  \]
  with $g_k(x) = \hat w_k^\top x + \hat b_k$, where 
  for $k =1, \ldots, d$,
   $(\hat f_k, \hat w_k, \hat b_k)$ solves the sequential optimization problem:
  \begin{mini}
    {f \in \mathcal{H}, w \in \mathbb R^p, b \in \mathbb R}
    {\sum_{i=1}^n \left( f(z_i) - w^\top x_i - b \right)^2 + \lambda_w \|w\|_2^2 + \lambda_f J(f) }
    {}
    {}
    \addConstraint{\sum_{i=1}^n f(z_i) = 0, \quad \frac{1}{n}\sum_{i=1}^n [f(z_i)]^2 = 1}{}
    \addConstraint{f \perp \hat f_{k-1}, \ldots, \hat f_1}{}
    \label{eq:probe_f}
  \end{mini}
  and $(\hat u_k, \hat c_k)$ solves the optimization problem:
  \begin{mini*}
    {u,c \in \mathbb R^p}
    {\sum_{i=1}^n \left\| x_i - u f_k(z_i) - c \right\|^2.}
    {}
    {}
  \end{mini*}
\end{definition}

For fixed regularization parameters $\lambda_w, \lambda_f$, the Manifold Probe has a closed-form solution. 

We'll define the centered model matrices $X \in \mathbb R^{n \times p}$ and $H \in \mathbb R^{n \times m}$ with rows $X_{i,:} = x_i - \bar x$ and $H_{i,:} = h(z_i) - \bar h$ respectively, where $\bar x = (1/n)\sum_{i=1}^n x_i$ and $\bar h = (1/n)\sum_{i=1}^n h(z_i)$.

The closed-form solution that we present requires that the matrix $H$ has full-column rank, so that all of the coefficients $\beta$ can be uniquely estimated. If this is not the case (which is likely given the centering), we can linearly reparametrize the basis so that it does\footnote{for example, using its singular value decomposition.}. From hereon, we will assume that the basis is parametrized such that $H$ has full-column rank.

\begin{proposition}
  \label{prop:closed_form_solution}
  Let $(\hat f_k, \hat w_k, \hat b_k)$, be the solutions to the optimization problem \eqref{eq:probe_f}. Then, $\hat f_k(z) = \hat \beta_k^\top h(z)$ where $\hat \beta_1,\ldots, \hat \beta_d$ are an orthonormal set of eigenvectors corresponding to the $d$ smallest eigenvalues $\hat \nu_m, \ldots, \hat \nu_{m-d}$ of the generalized eigenvalue problem
  \begin{equation*}
    \label{eq:eval_eq}
    M \beta = \nu \Sigma \beta
  \end{equation*}
  where
    \[
      M := H^\top(I - A)H + \lambda_f S, \qquad A = X(X^\top X +\lambda_w I)^{-1} X^\top, \qquad \Sigma = \frac{1}{n}H^\top H.
    \]
  In addition,
  \[
    \hat w_k = (X^\top X + \lambda_w I)^{-1} X^\top H \hat \beta_k, \qquad \hat b_k = -\hat w_k^\top \bar x, \qquad \hat u_k = \frac{1}{n} X^\top H \hat \beta_k
  \]
  where $\bar x = (1/n)\sum_{i=1}^n x_i$.
\end{proposition}

A proof of Proposition~\ref{prop:closed_form_solution} is given in Section~\ref{sec:proof_of_closed_form_solution} of the appendix.

\subsection{Fitting the regularization parameters}

While Proposition~\ref{prop:closed_form_solution} tells us how to fit the Manifold Probe for a fixed pair of regularization parameters $\lambda_w, \lambda_f$, it doesn't tell us anything about how we should choose them. In practice, we will want to choose them using the data, and we will want to choose different regularization parameters for each sequentially fitted feature.

One approach is to directly apply $k$-fold cross-validation to the objective, searching over candidate parameter values using, for example, grid search.

In Section~\ref{sec:selecting_regularization_parameters} of the appendix, we present an alternative approach which we favour in practice. We show that \eqref{eq:probe_f} can be optimized by solving a sequence of alternating (generalized) ridge regression problems,
 which we prove converge to the global minimizer under very mild conditions. 
We then propose to select the regularization parameters at each iteration using a closed-form criterion appropriate for ridge regression. In practice, we use either Generalized Cross-Validation \citep{craven1978smoothing, wood2004stable} or Restricted Maximum Likelihood \citep{bartlett1937properties, wood2011fast}, which have closed-forms and can be optimized very quickly using Newton's method, without refitting the model. 

\section{Discovering time and space manifolds in large language models}
\label{sec:discovering}

In this section, we use the Manifold Probe to discover hidden time and space manifolds in the residual stream of Llama 2-7b \citep{touvron2023llama}, an open-weights large language model. 
We demonstrate how we can use the probe both as an interpretability tool, to discover features which are linearly represented in the residual stream; and as a steering tool, to causally influence the model's behaviour.

\begin{figure}[t]
  \centering
  
  \includegraphics[height=1.4em]{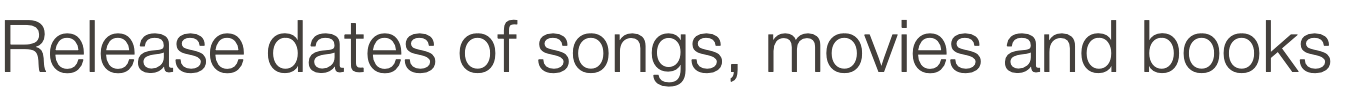}
  \vspace{1em}
  
  \includegraphics[width=0.35\textwidth, trim=1.5cm 0 0 3.7cm, clip]{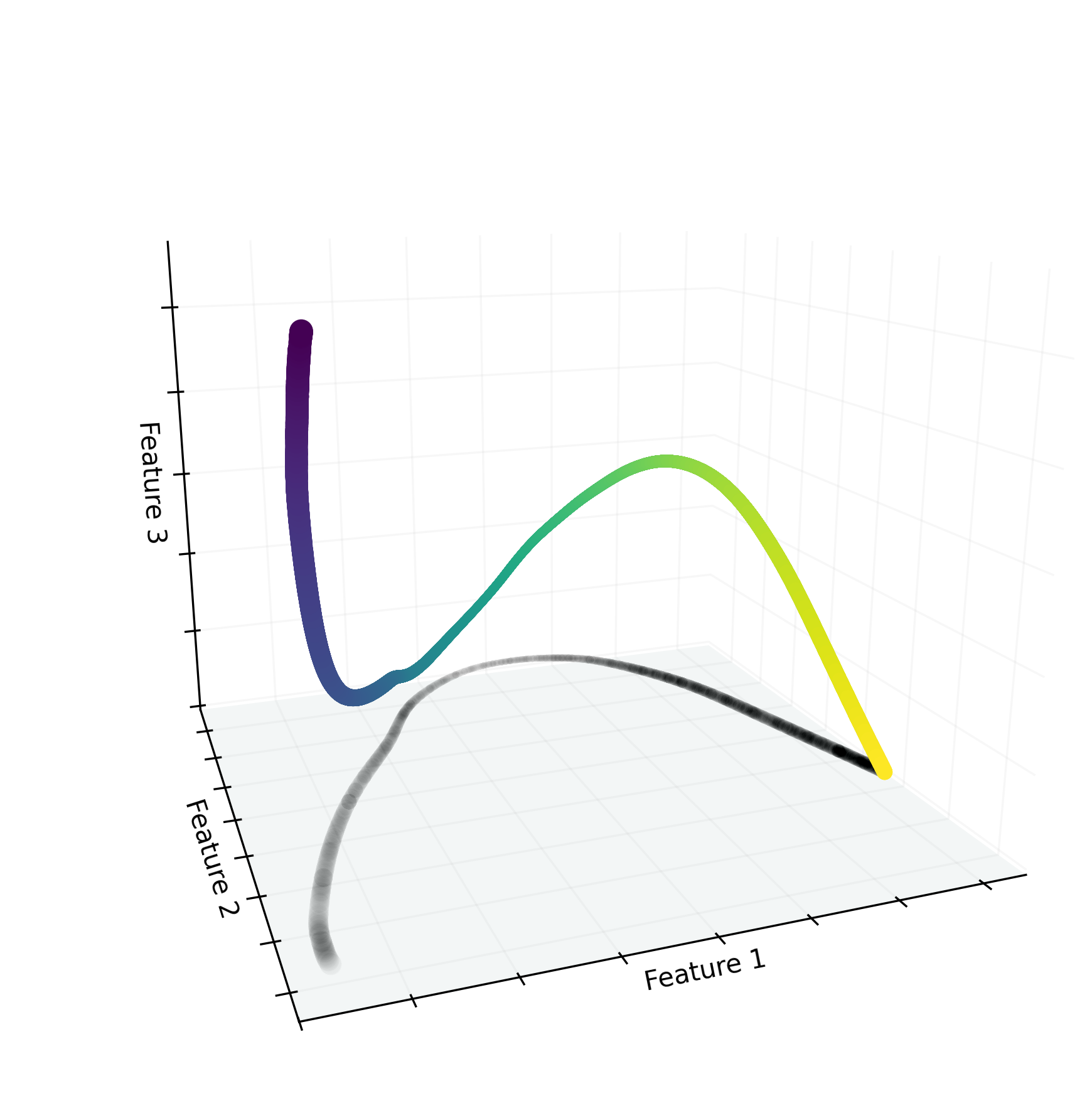}
  \includegraphics[width=0.35\textwidth, trim=1.5cm 0 0 3.7cm, clip]{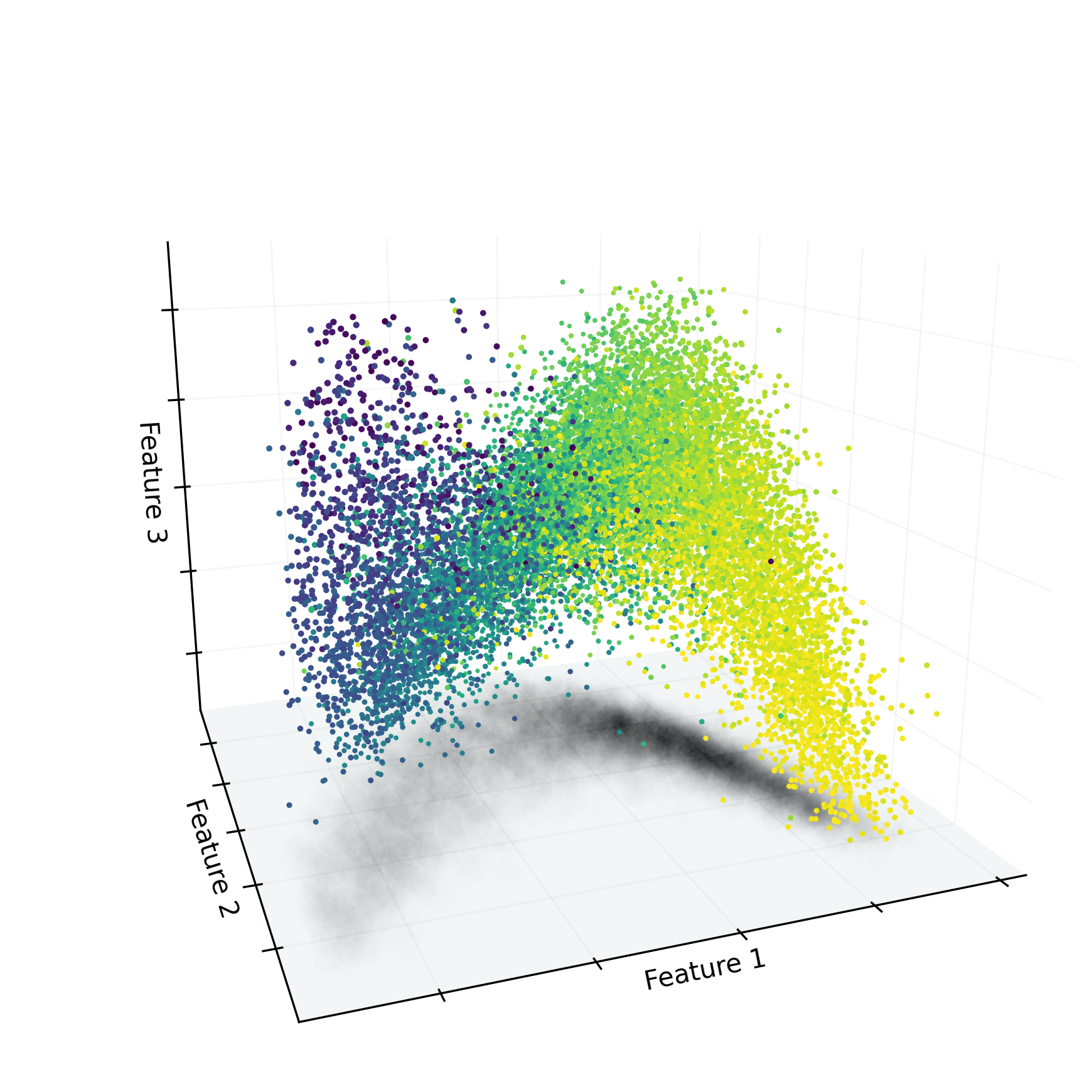}
  
  \includegraphics[width=1\textwidth, trim=0 0 0 0, clip]{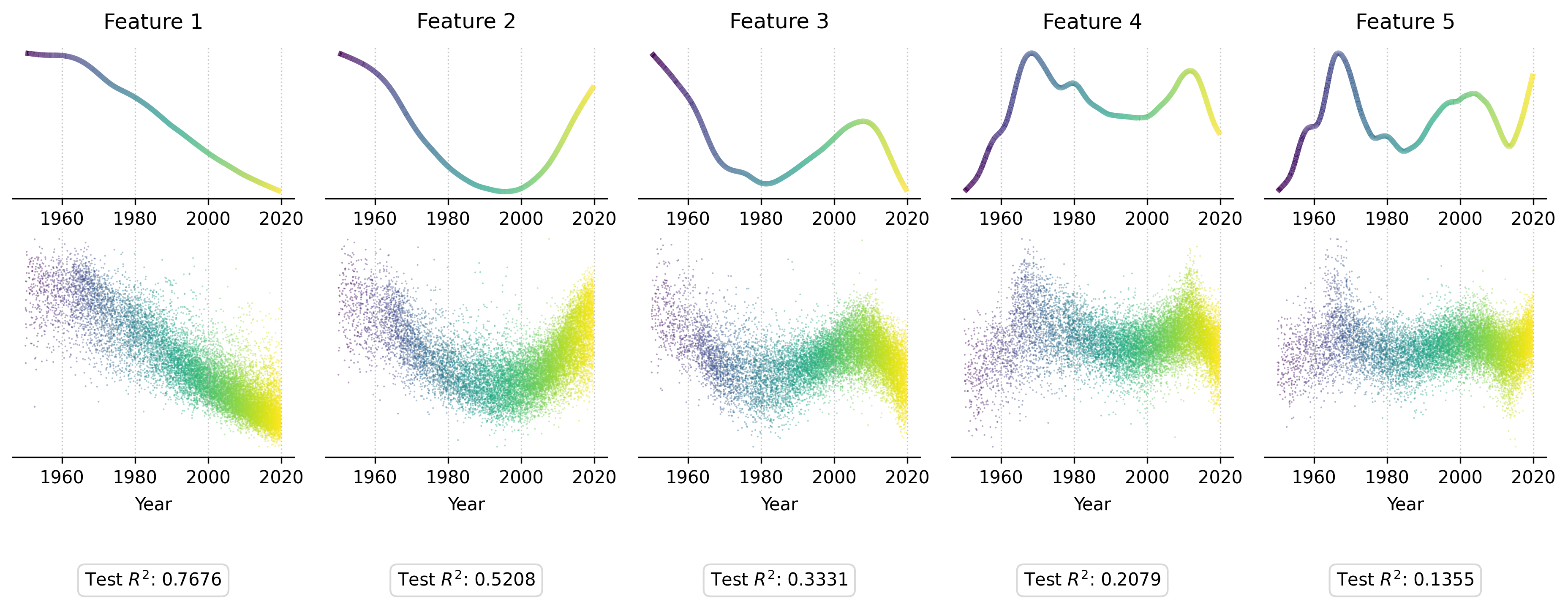}
  \caption{
    A representation manifold (top left) and linear prediction (top right) from a Manifold Probe fitted the release dates of songs, books and movies from layer 16 residual stream activations of Llama 2-7b. \emph{Below:} the first five fitted features (top row), corresponding linear predictions (bottom row) for representations in the test set, and test $R^2$ coefficients.}
    \label{fig:time_manifold}
  \end{figure}

We make use of two probing datasets collected by \citet{gurnee2024language}. The first dataset contains the names and creators of popular songs, movies and books alongside their corresponding release dates (represented as a decimal year); and the second contains the names and geographic coordinates of places in the U.S.A.. After some filtering, we have 29,503 works released in $\*Z_{\texttt{time}} = [1950, 2020]$, and 17,381 places with coordinates in $\*Z_{\texttt{space}} = [24.5, 49.5] \times [-125.0, -66.5]$ (the bounding box of mainland U.S.A.). In both cases, we consider a 50-50 train/test split.

\begin{figure}[t]
  \centering
  
  \includegraphics[height=1.3em]{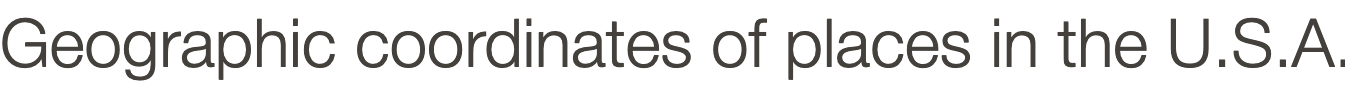}
  \vspace{1em}
  
  \includegraphics[width=0.35\textwidth, trim=1.5cm 0 0 3cm, clip]{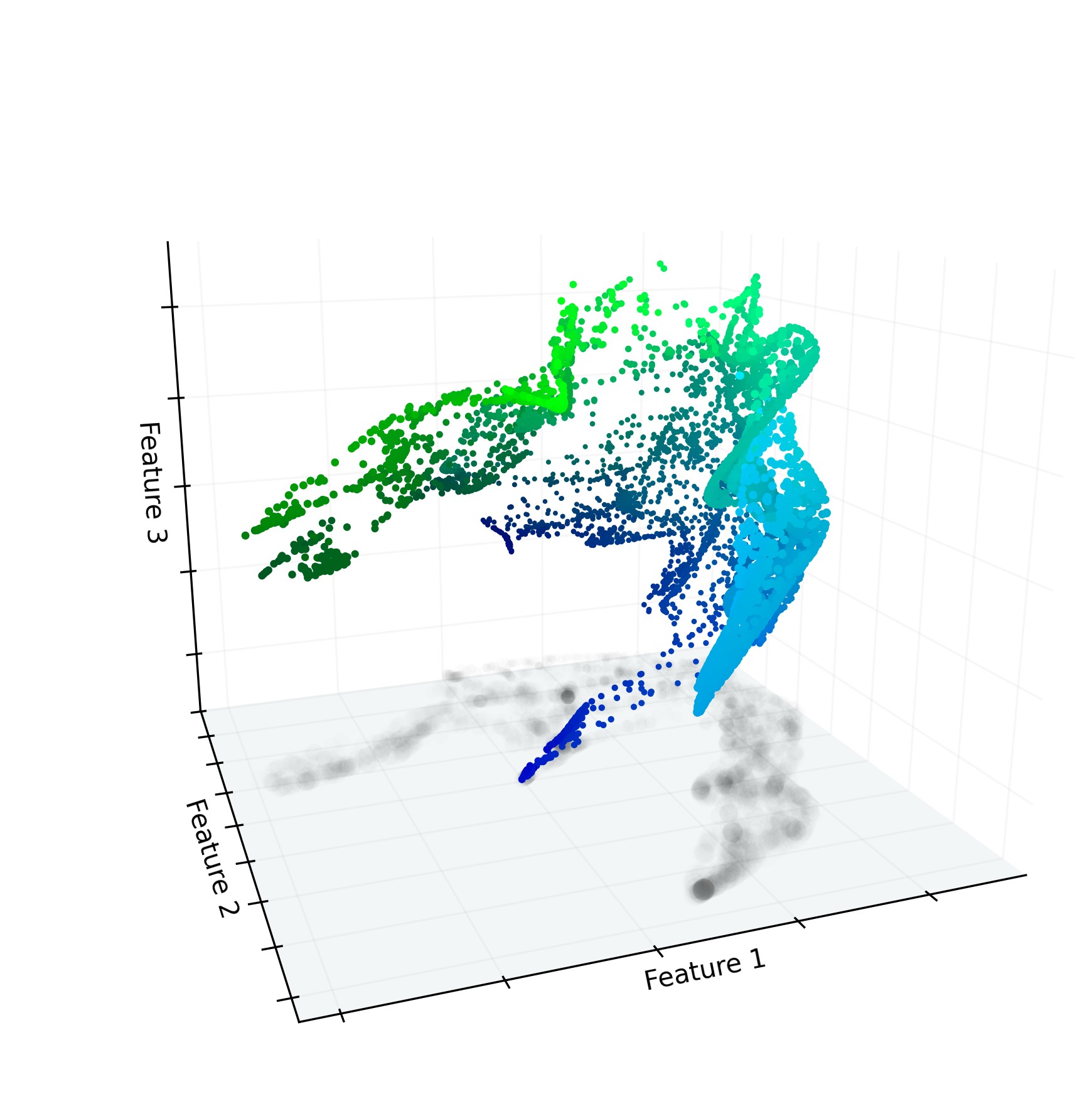}
  \includegraphics[width=0.35\textwidth, trim=1.5cm 0 0 3cm, clip]{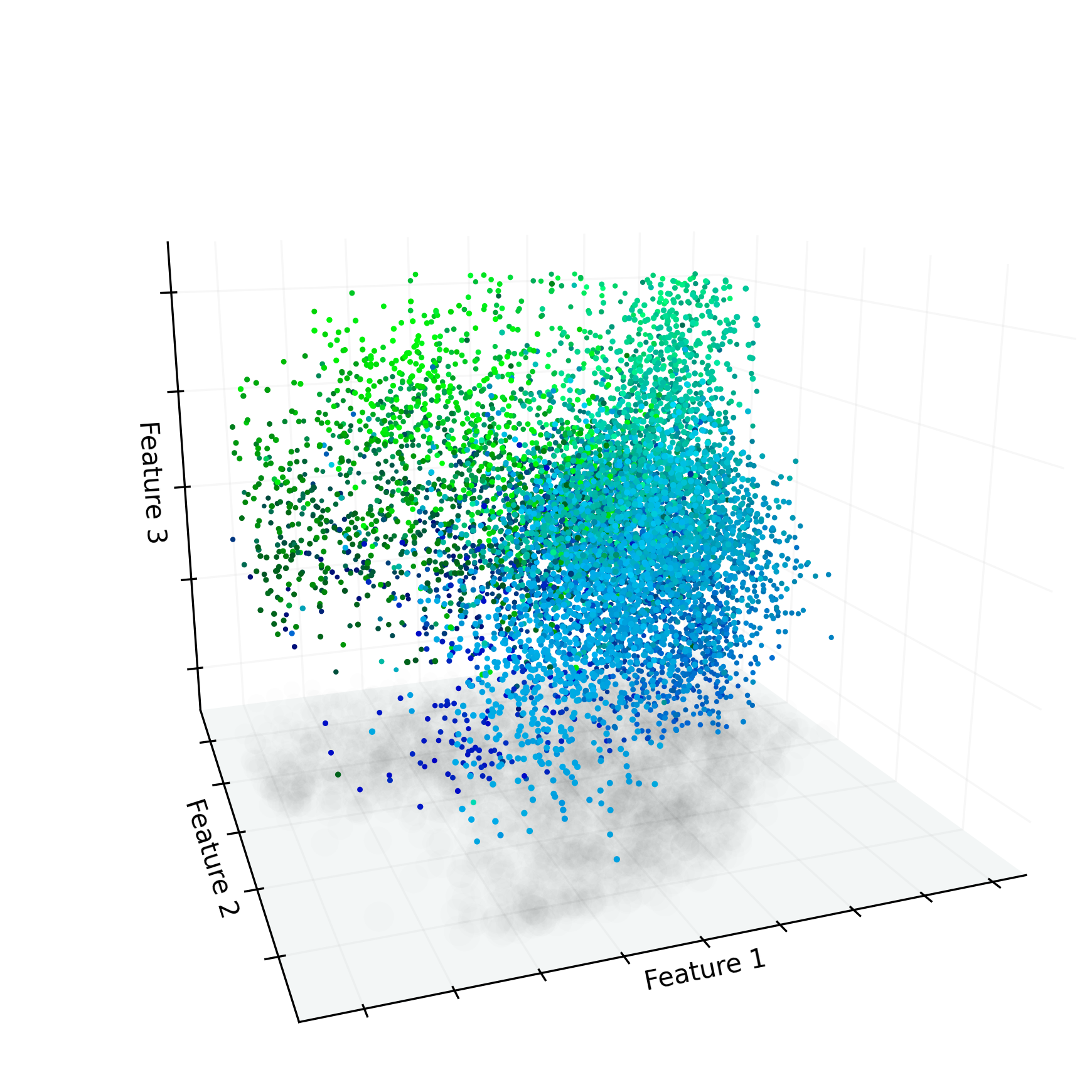}
  
  \includegraphics[width=0.3\textwidth, trim=0 0 0 0, clip]{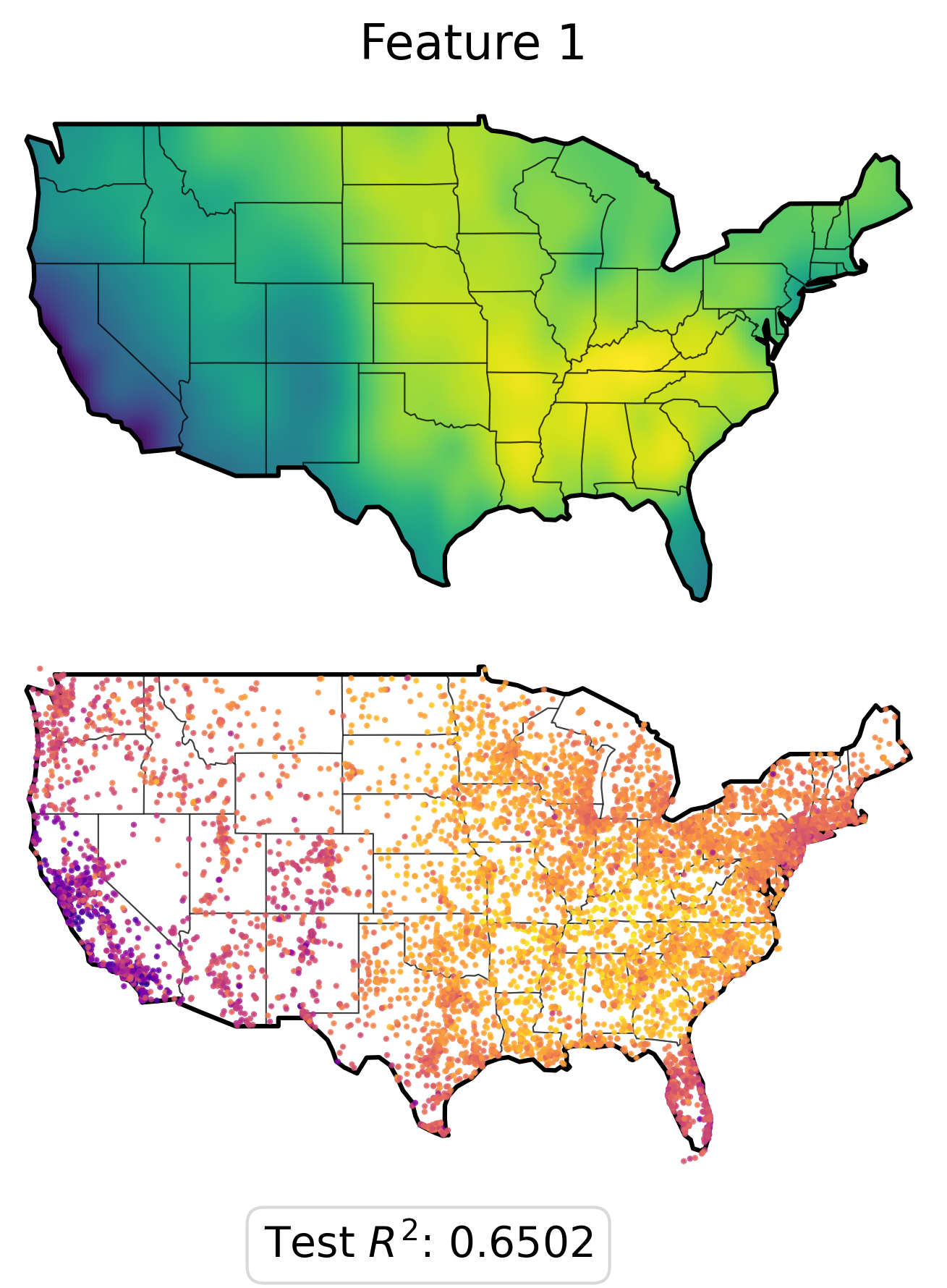}
  \includegraphics[width=0.3\textwidth, trim=0 0 0 0, clip]{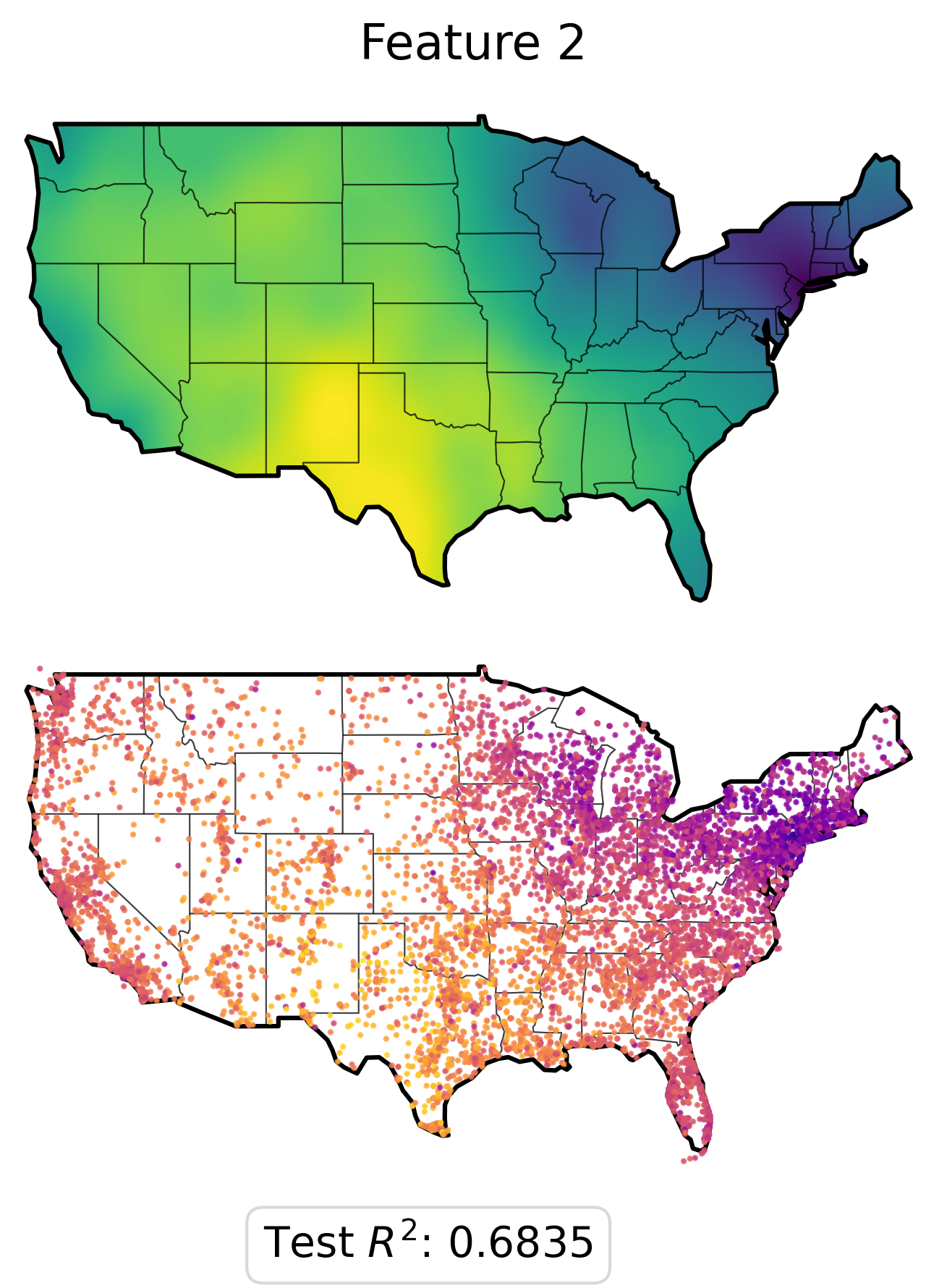}
  \includegraphics[width=0.3\textwidth, trim=0 0 0 0, clip]{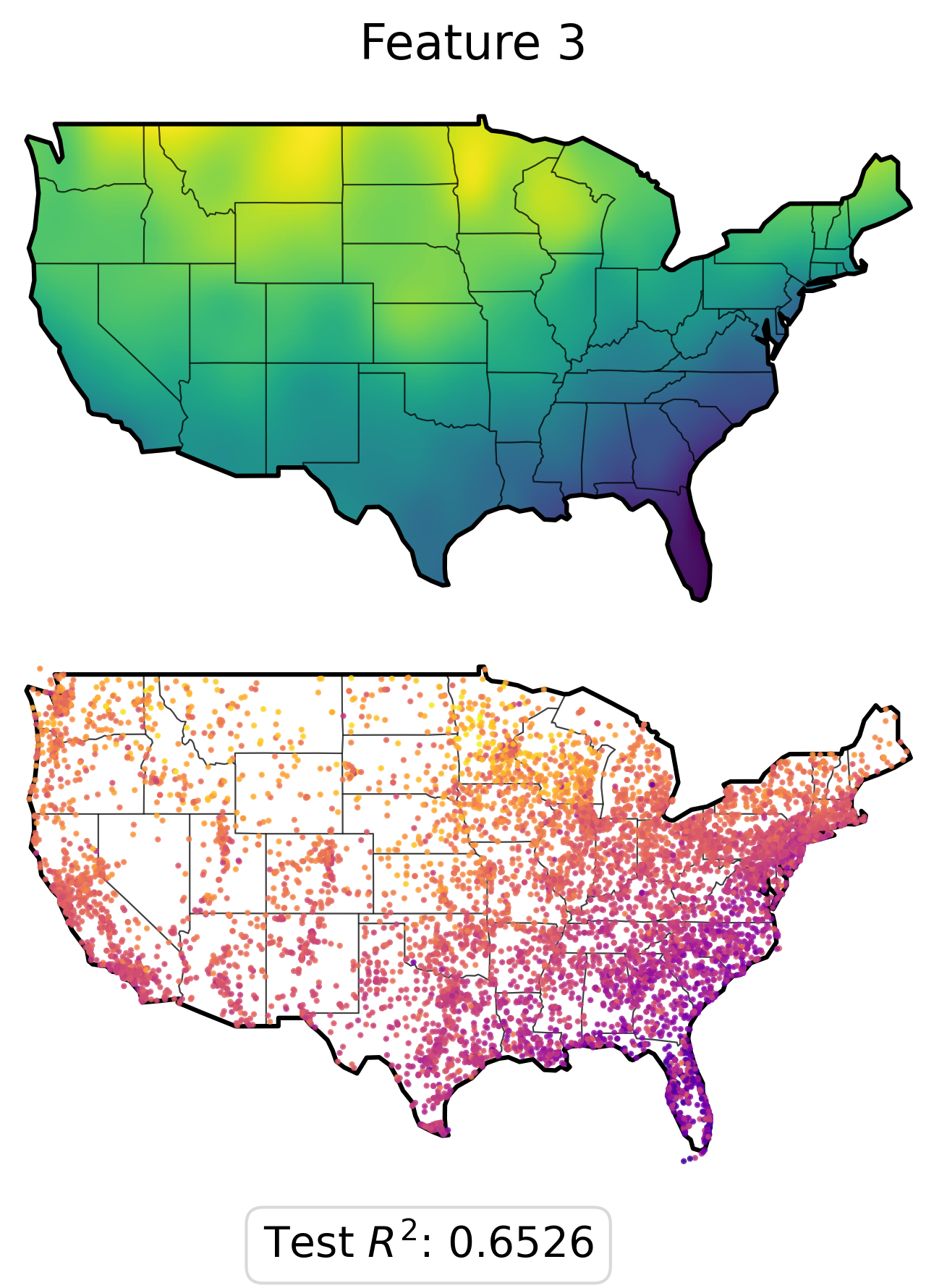}
  \caption{A representation manifold (top left) and linear prediction (top right) from a Manifold Probe fitted to the geographic coordinates of places in the U.S.A. from layer 16 residual stream activations of Llama 2-7b. \emph{Below:} the first three fitted features (top row), corresponding linear predictions (bottom row) for representations in the test set, and test $R^2$ coefficients. Feature values are given by colour.
  }
  \label{fig:place_manifold}
\end{figure}

For each entity, we construct a string such as ``Queen's Bohemian Rhapsody'' or ``Lake of the Ozarks'' which we feed into the language model, and record the last token residual stream activations at each layer. To train the manifold probe, we parametrize time features using cubic B-splines with 280 knots, and parametrize space features using a tensor product of two cubic B-splines with 40 and 80 knots respectively. We use the fitting procedure described in Section~\ref{sec:selecting_regularization_parameters}, and use the REML criterion to select the regularization parameters.

\subsection{Interpretability: exploring linearly-represented features}

The bottom row of plots of Figures~\ref{fig:time_manifold} and \ref{fig:place_manifold} show the first few features $\hat f_1,\hat f_2,\ldots$ fitted by the probe to the layer 16 representations from the two training sets, and the corresponding linear predictions $g_1(x), g_2(x), \ldots$ of these features from representations $x$ in the test sets. We report the $R^2$ coefficients of these test predictions, which measure the extent to which the features are linearly represented. Perfect predictions have an $R^2$ coefficient of one, predicting the feature mean always has an $R^2$ coefficient of zero, and predictions which are worse than predicting the mean have a negative $R^2$ coefficient. The top left plots show three dimensions of the estimated manifolds $\hat{\*M}$ and the manifold predictions $\Psi(x)$ from the representations $x$ in the test set, with respect to the first three fitted basis vectors $\hat u_1, \hat u_2, \hat u_3$.

Since the features $f_1,\ldots,f_d$ in the decomposition \eqref{eq:decomposition} are only defined up to rotations of the basis $u_1, \ldots, u_d$, it can be informative to apply factor analysis to the learned features to rotate them into a basis in which they are more easily interpretable. We take the top 5 time features, and the top 32 space features, and apply Varimax rotation \citep{kaiser1958varimax, rohe_vintage_2023} which aims to make the rotated features approximately sparse. The resulting features are shown in Figure~\ref{fig:varimax} in the appendix. It is of particular note that rotated space features localize on many U.S. states, showing that they are approximately linearly separated in the representations. We can also interpret from the rotated time features that the decades from the 1950s to the 2010s are approximately linearly separated.

\begin{figure}[t]
  
  \centering

  \begin{minipage}{0.49\textwidth}
    \centering
    \vspace{-0.1em}
    \hspace{-1.7em}
    \includegraphics[height=1em]{imgs/art_title.png}\\[-0.1em]
    \includegraphics[width=\textwidth, trim=0.7em 0.5em 0.7em 0.8em, clip]{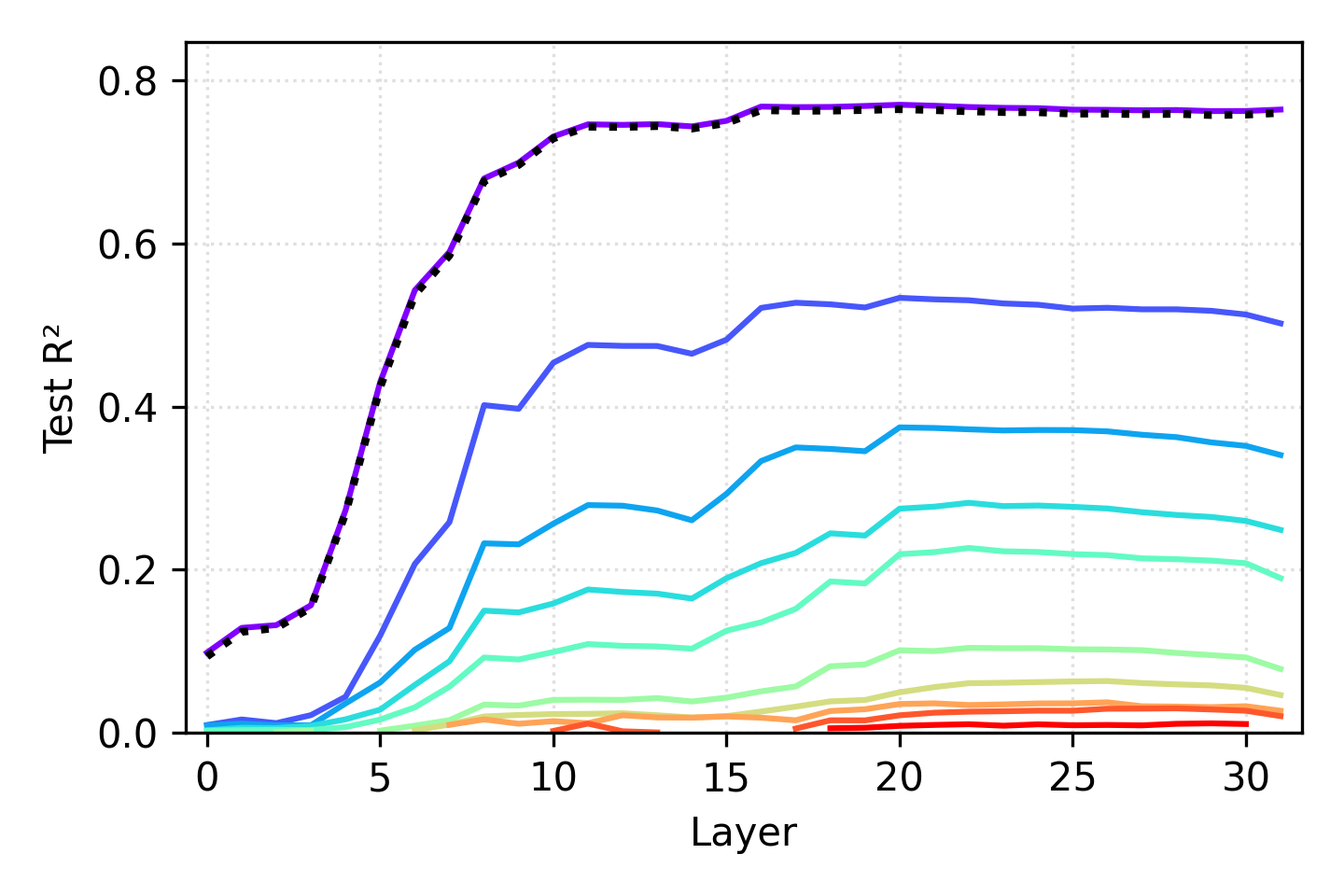}
  \end{minipage}
  \hfill 
  \begin{minipage}{0.49\textwidth}
    \centering
    \hspace{-1.7em}
    \includegraphics[height=1em]{imgs/us_place_title.png}\\
    \includegraphics[width=\textwidth, trim=0.7em 0 0.7em 0.6em, clip]{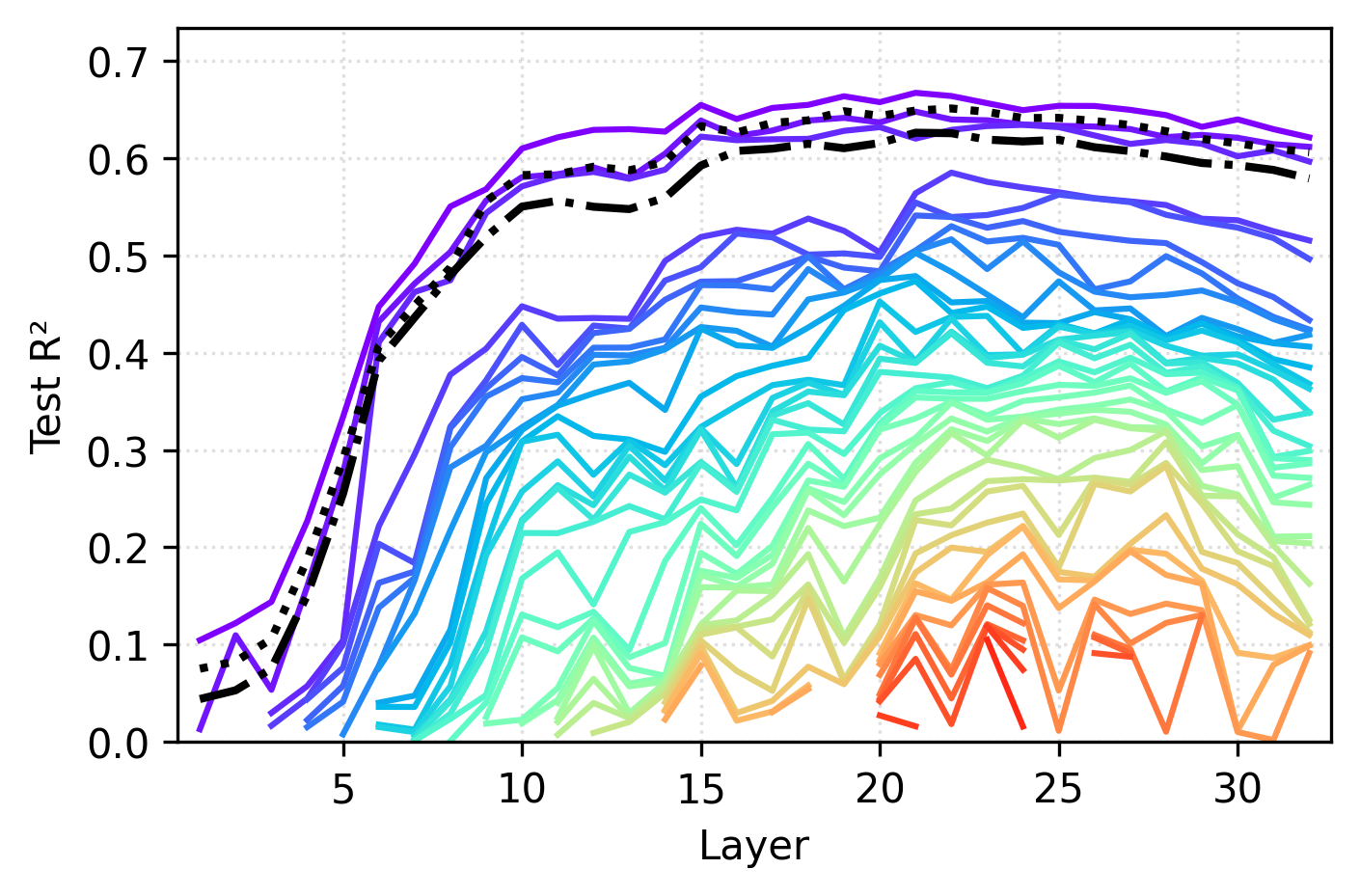}
  \end{minipage}

  \vspace{-0.5em}

  \includegraphics[width=0.4\textwidth]{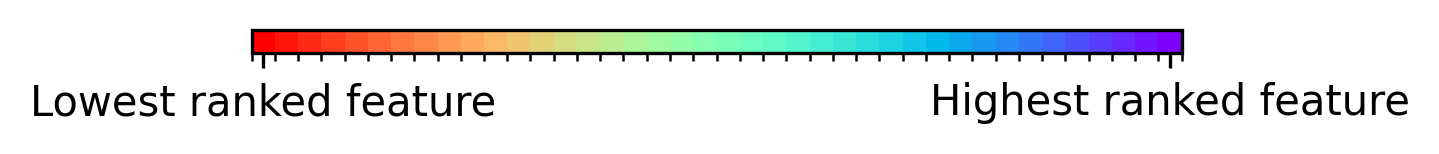}

  \caption{Ranked test $R^2$ values for features fitted using the probing datasets described in Section~\ref{sec:discovering} at each layer of Llama 2-7b. \emph{Left:} the dotted line shows the test $R^2$ coefficient of a ridge regression fit directly to the release dates from the songs, movies and books representations. \emph{Right:} the dotted and dashed lines show the test $R^2$ coefficients of ridge regression fits directly to the latitude and longitude from the U.S. places representations respectively.}
  \label{fig:r2_lines}
\end{figure}

To get an idea for how much information about the release dates and locations is linearly represented in the residual stream at each layer, we fitted features until the corresponding test $R^2$ coefficients were continually below zero. In Figure~\ref{fig:r2_lines}, we plot the value of the $k$th ranked test $R^2$ coefficient for each layer where this is above zero. In both datasets, the predictabilities of features increase in predictability throughout the first half of the layers before levelling off. The location representations consistently contain three features which are considerably more predictable that the rest. 

We include the test $R^2$ coefficients of a ridge regression fit directly to the release dates (dotted line) from the songs, movies and books representations; and to the latitude (dotted line) and longitude (dashed line) from the U.S. places representations. These were reported in \citet{gurnee2024language} as evidence that language models linearly represent space and time. 
In the time representations, we find that for all layers the most linearly predictable feature is very close to the identity feature, and so the test $R^2$ coefficient for our highest ranked feature and the direct year track very closely. In the time representations, the highest ranked feature we find has a higher test $R^2$ coefficient than the latitude and longitude features.

\subsection{Steering: causally influencing the model's understanding of time}

In this section, we demonstrate that we can use the representation manifold learned by the Manifold Probe to causally influence the model's internal belief about the release dates of songs, movies and books.

To do this, we treat the learned concept representations $\hat \phi(z)$ as an infinite continuum of steering vectors which trace the manifold $\hat{\*M}$. To steer the model's internal belief about the value of a concept to a particular value $z$, we propose to intervene on a representation $x$ by adding a scalar multiple of $\phi(z)$ to it, i.e. setting
\[
  x \longleftarrow x + \alpha \hat \phi(z)
\]
for some $\alpha > 0$.

From the probing dataset of songs, movies and books, we fit three-dimensional manifolds $\hat \phi_l$ to the last-token residual stream activations at each layer of the model as in the previous section. We then select a stratified sample of 1,400 works from the test set, with 200 from each decade, to use for our steering experiment. For each song, movie, or book, we construct a prompt of the form
\[
  \text{\textit{``}$\langle$creator$\rangle$\textit{'s} $\langle$title$\rangle$ \textit{was released in the year}''}
\]
which we feed into the model. Assuming standard temperature-one sampling, we record the probability of the model completing the prompt each year between 1945 and 2025.

\begin{figure}[t]
  \centering
  
  \includegraphics[width=\textwidth]{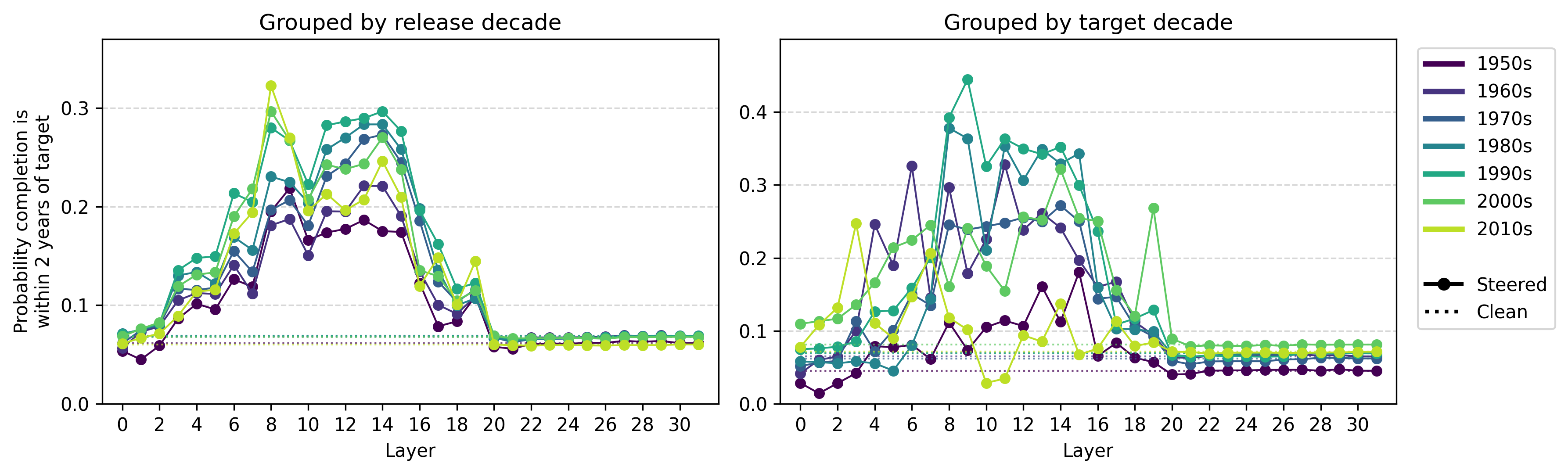}
  
  \includegraphics[width=\textwidth]{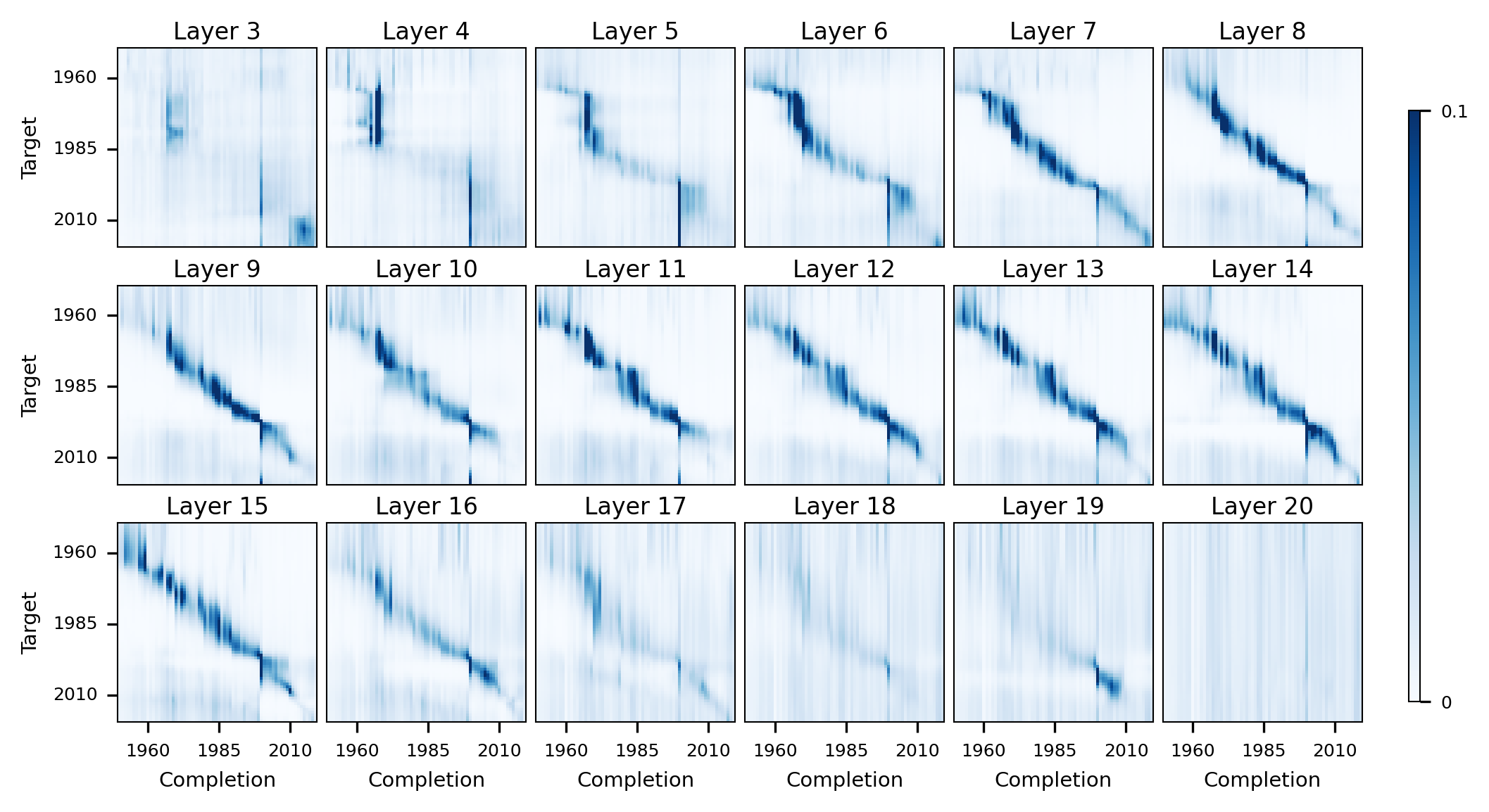}
  
  \caption{Steering experiment. \emph{Top:} the mean probability a completion is within two years of the target year it was steered to at each layer, grouped by release decade (left) and target decade (right). Clean baselines are shown with dashed lines. \emph{Bottom:} colour intensity (capped at 0.1) indicates the mean probability of a completion given the steering target.}
  \label{fig:steering}
\end{figure}

We do this for a clean run, and for each layer $l$ and year $z$ between 1950 and 2020, we do this again but intervene by steering the residual stream activations of the last token of the work's title at layer $l$ by the steering vector $\hat \phi_l(z)$ and $\alpha=100$. This results in a total of $32 \times 70 = 2240$ interventions per prompt\footnote{The full experiment took approximately 100 GPU hours on Nvidia RTX 4090s.}. To measure efficacy of an intervention, we report the probability that the model completes the prompt with a year that is within two years of the target $z$.

The top plots in Figure~\ref{fig:steering} show the mean efficacy of the interventions for each layer, grouped in the left-hand plot by the decade of the song, movie or book, and in the right-hand plot by the decade of the target year $z$. From the left-hand plot, we see that the efficacy of the interventions peak at layers 8 and 14, depending on the release decade of the work, and that this drops sharply after layer 15, and dropping to baseline performance at layer 20 and beyond. From the right-hand plot, we see that steering efficacy and the optimal layer for intervention depends quite heavily on the target year.

Figure~\ref{fig:degradation} in the appendix shows the probability that the model completes the prompt with a year at all. With the exception of steering to years in the 1950s, we see that these interventions have very little effect on the model's ability to meaningfully complete the prompt.

For layers 3 to 20, the bottom plot in Figure~\ref{fig:steering} shows the mean probability of each completion for each steering target. The dark diagonal streaks indicate that the interventions are having some success at influencing the model to complete the prompt with a given target year. Figure~\ref{fig:steering_std} in the appendix shows the standard deviations.

\section{Discussion}

In this work, we introduced the Manifold Probe, a supervised method for discovering representation manifolds in superposition.
We hope that our probe will serve as a useful new tool for the mechanistic interpretability community. We see potential applications as a data-driven way to discover mechanisms of continuous computation such as counting \citep{gurnee2025when, wu2025uncovering} and modular arithmetic \citep{nanda_progress_2023,zhong_clock_2023}, and as a tool to map out abstract concepts such as emotion \citep{sofroniew2026twheemotion,sun2026valence}.
There are also potential implications for scientific discovery, for example, in interpreting recently-discovered phylogenetic and hematopoietic representation manifolds in biological foundation models \citep{pearce2025tree, wu2025uncovering}.

One limitation of our framework is that it implicitly assumes that the number of training samples is large relative to the dimension of the representation space, so that the sample estimates concentrate around their population counterparts. For state-of-the-art foundation models, this might require in the order of tens of thousands of examples. With smaller probing datasets, for good statistical performance it may be necessary to perform a preliminary principal component 
analysis to reduce the dimension of the representation space before fitting the probe.

Finally, while interpretability tools such as this one might be used to develop mechanistic guardrails or to steer model behaviour to serve the goals of safety, alignment and security, they might be used to learn to bypass safety guardrails or produce intentionally harmful behaviour, which as a community, we must be mindful of as we progress our scientific understanding of AI systems.

\section*{Acknowledgements}
The author would like to thank Jacob Davies, Jake Yukich, Can Rager, David Chanin, Nathalie Kirch, Patrick Rubin-Delanchy and Nick Whiteley for enlightening discussions on the topics of this paper.

\bibliographystyle{apalike}
\bibliography{bibliography.bib}


\newpage

\appendix

\section{Additional figures}
\label{sec:additional_figures}

\begin{figure}[H]

  \centering
  
  \includegraphics[height=1.3em]{imgs/art_title.png}

  \vspace{1em}

  \includegraphics[width=0.19\textwidth]{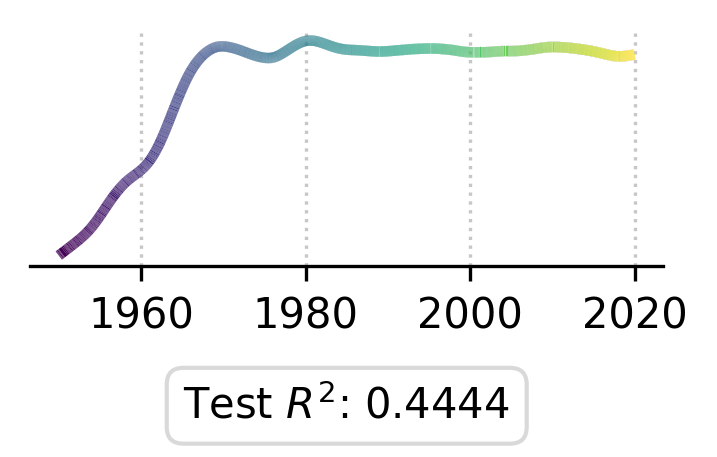}
  \includegraphics[width=0.19\textwidth]{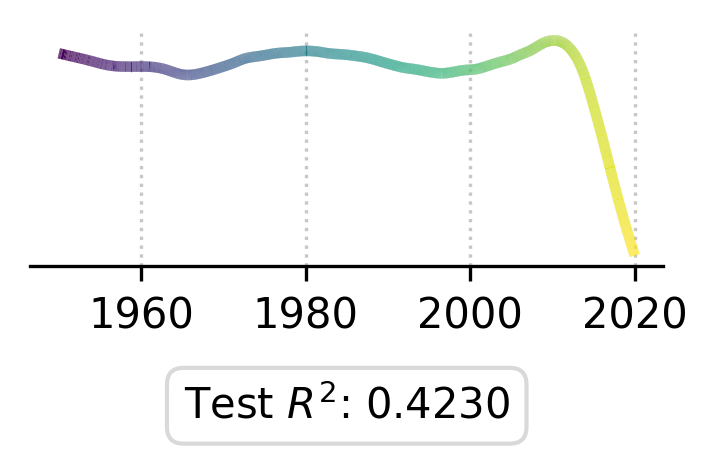}
  \includegraphics[width=0.19\textwidth]{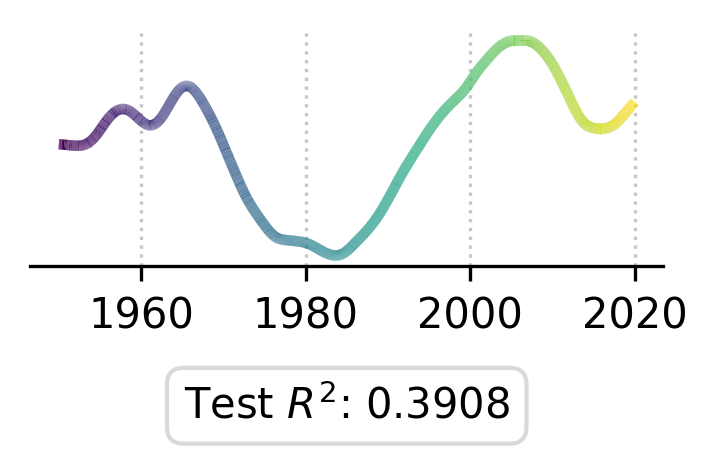}
  \includegraphics[width=0.19\textwidth]{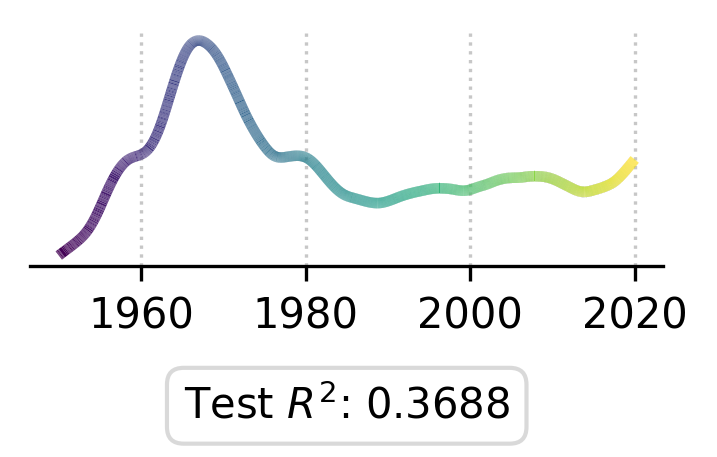}
  \includegraphics[width=0.19\textwidth]{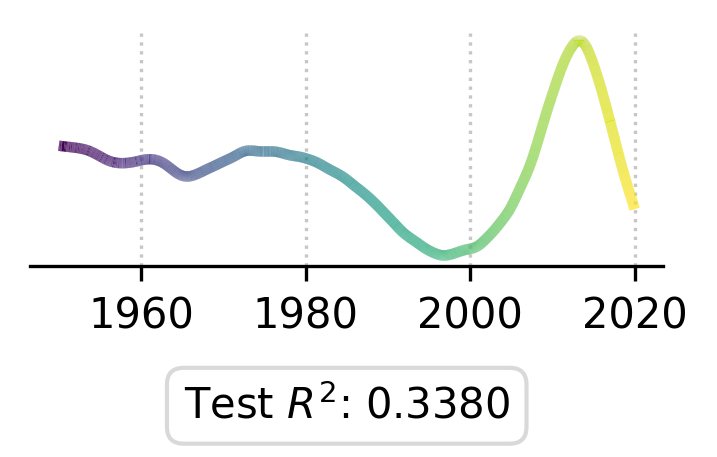}

\vspace{2em} 

  \includegraphics[height=1.3em]{imgs/us_place_title.png}
  
  \vspace{1em} 

  \includegraphics[width=0.23\textwidth]{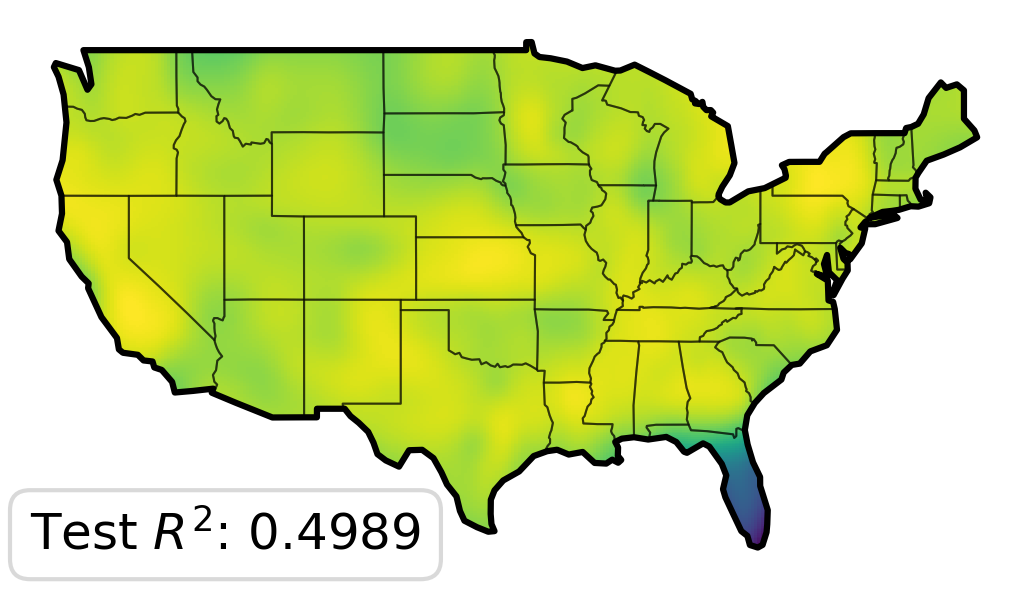}
  \includegraphics[width=0.23\textwidth]{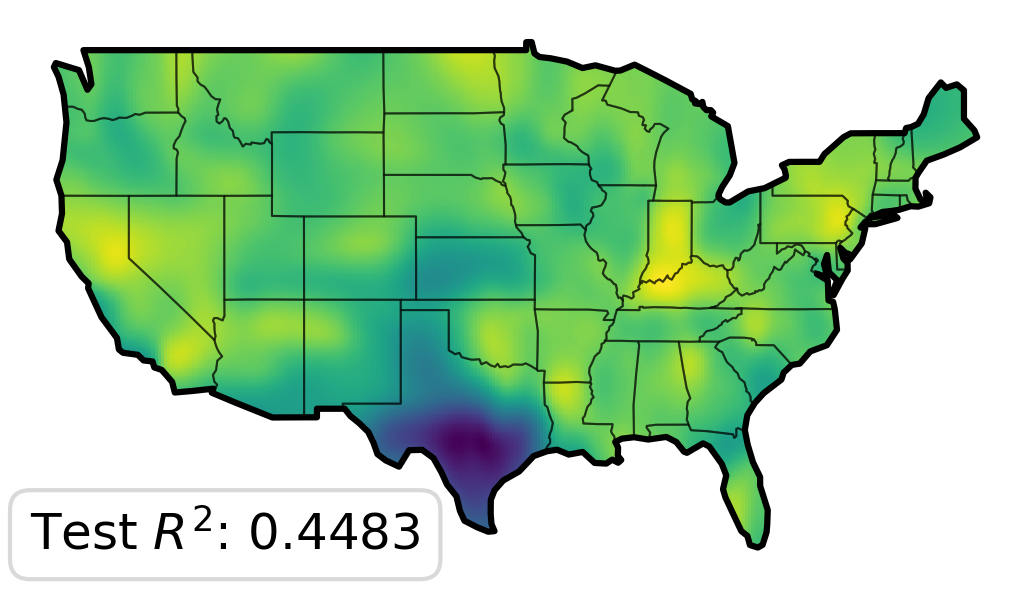}
  \includegraphics[width=0.23\textwidth]{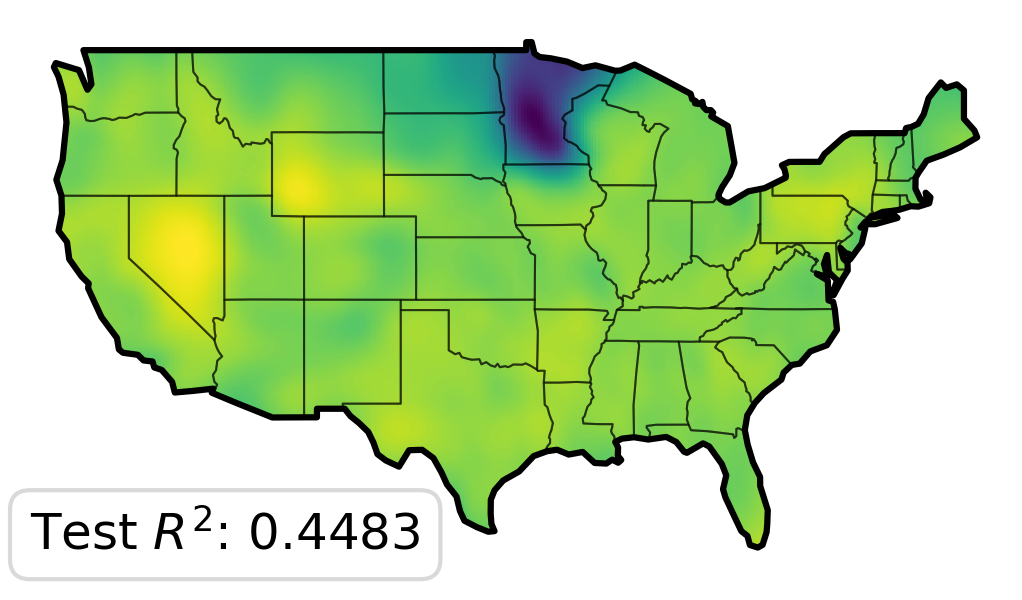}
  \includegraphics[width=0.23\textwidth]{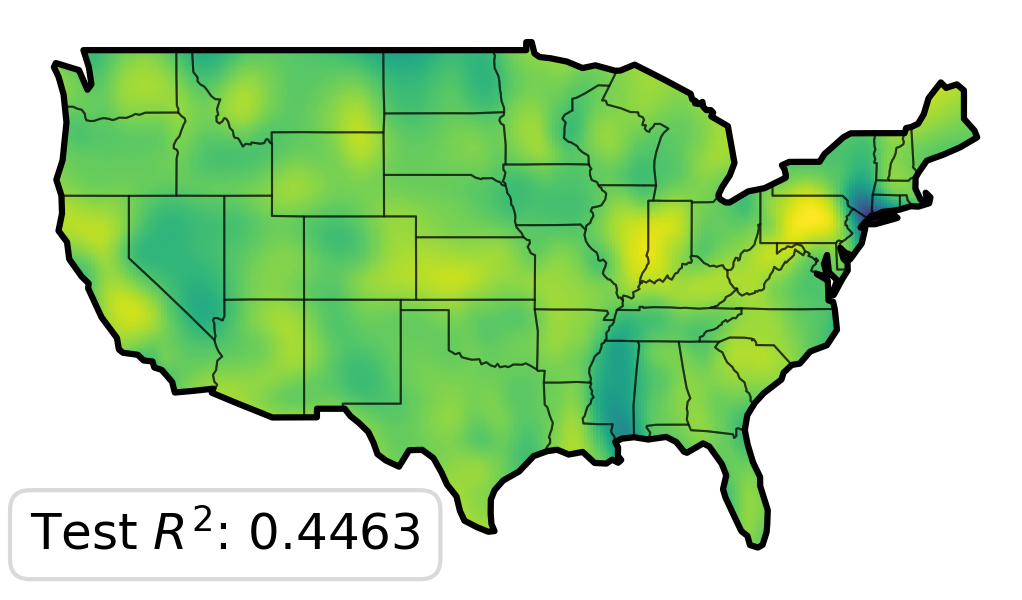}
  \includegraphics[width=0.23\textwidth]{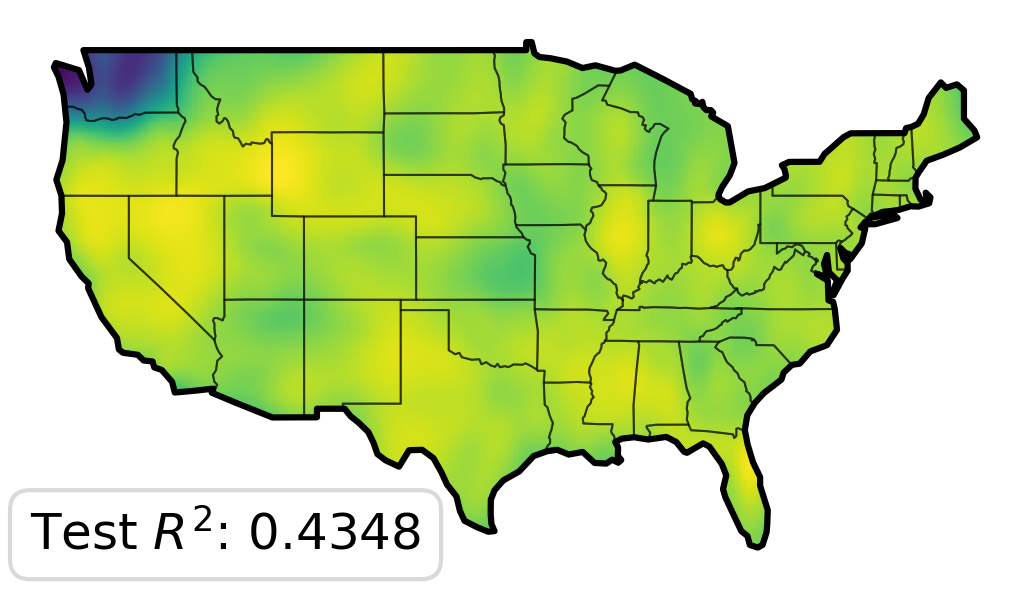}
  \includegraphics[width=0.23\textwidth]{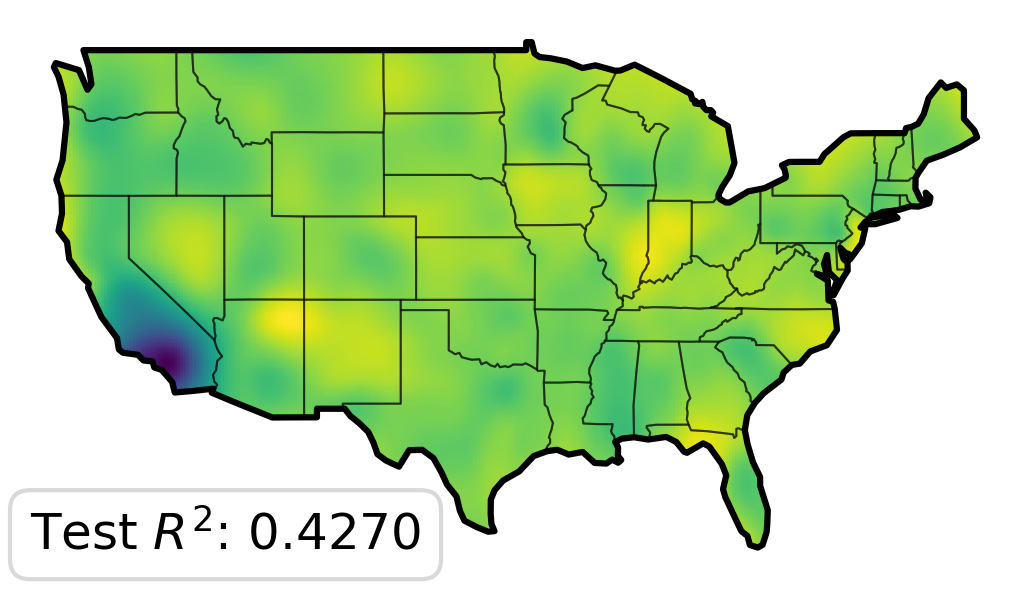}
  \includegraphics[width=0.23\textwidth]{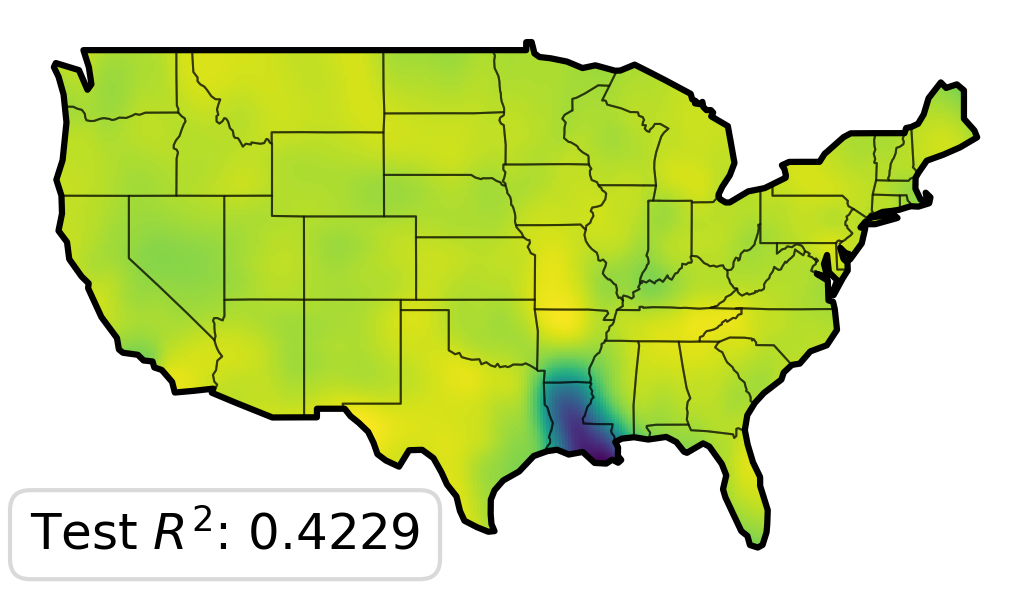}
  \includegraphics[width=0.23\textwidth]{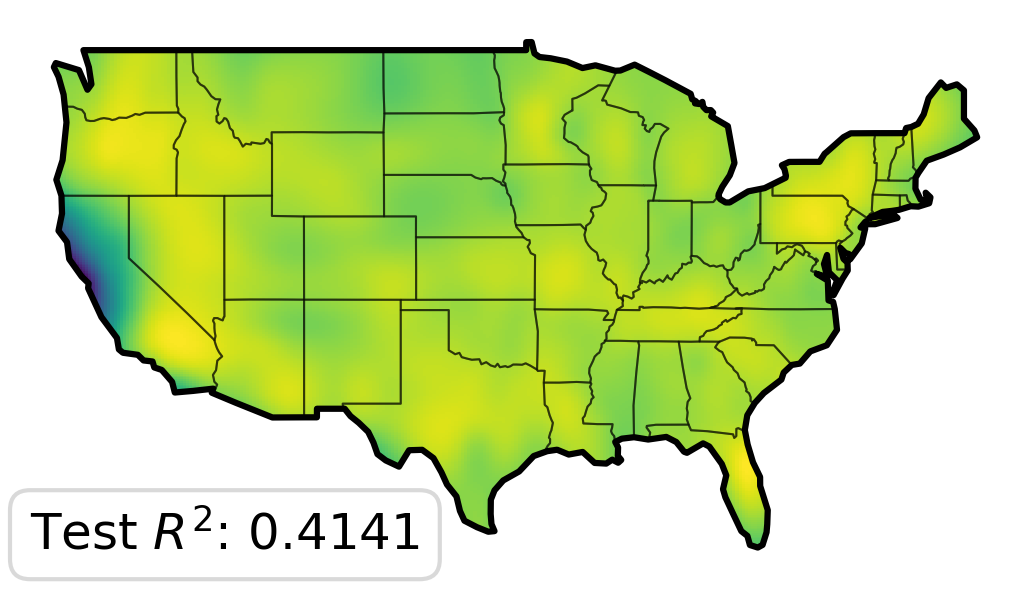}
  \includegraphics[width=0.23\textwidth]{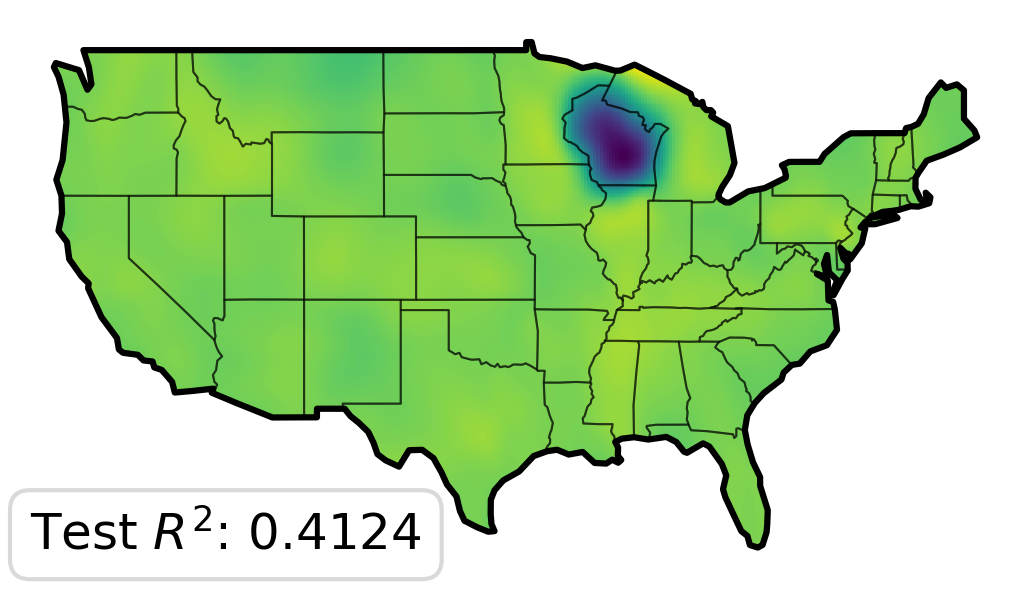}
  \includegraphics[width=0.23\textwidth]{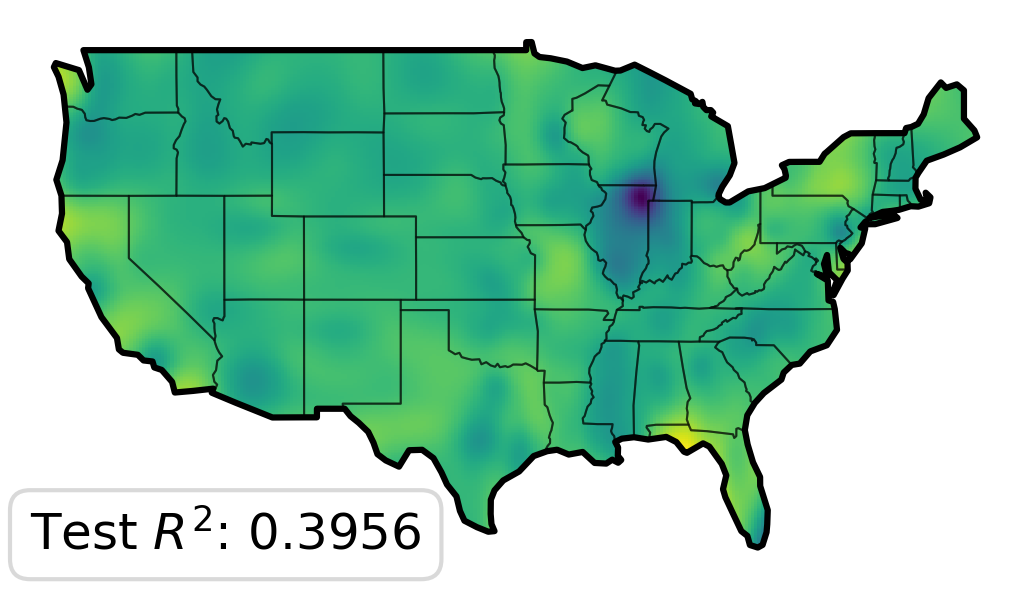}
  \includegraphics[width=0.23\textwidth]{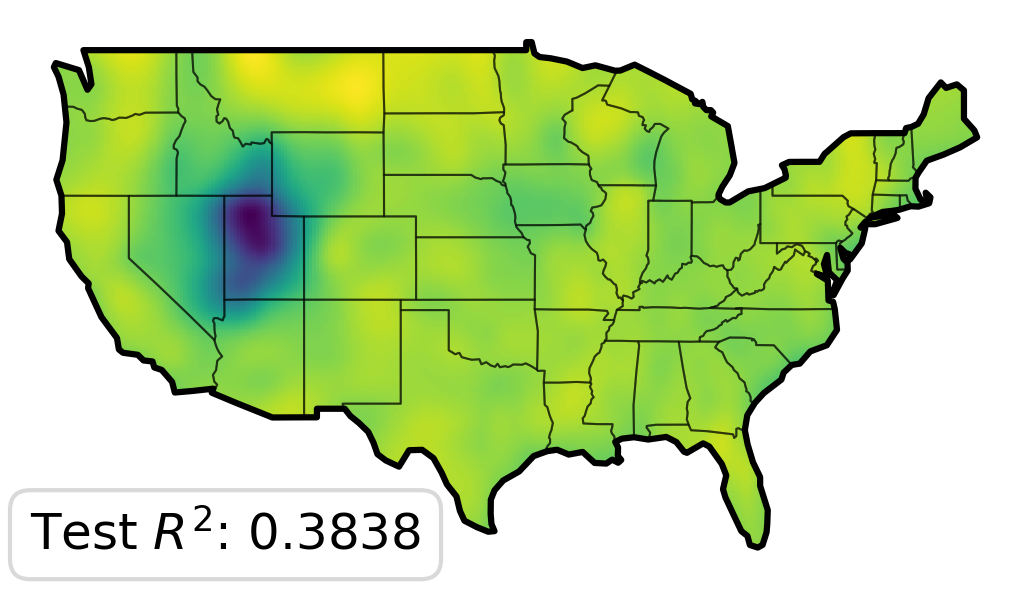}
  \includegraphics[width=0.23\textwidth]{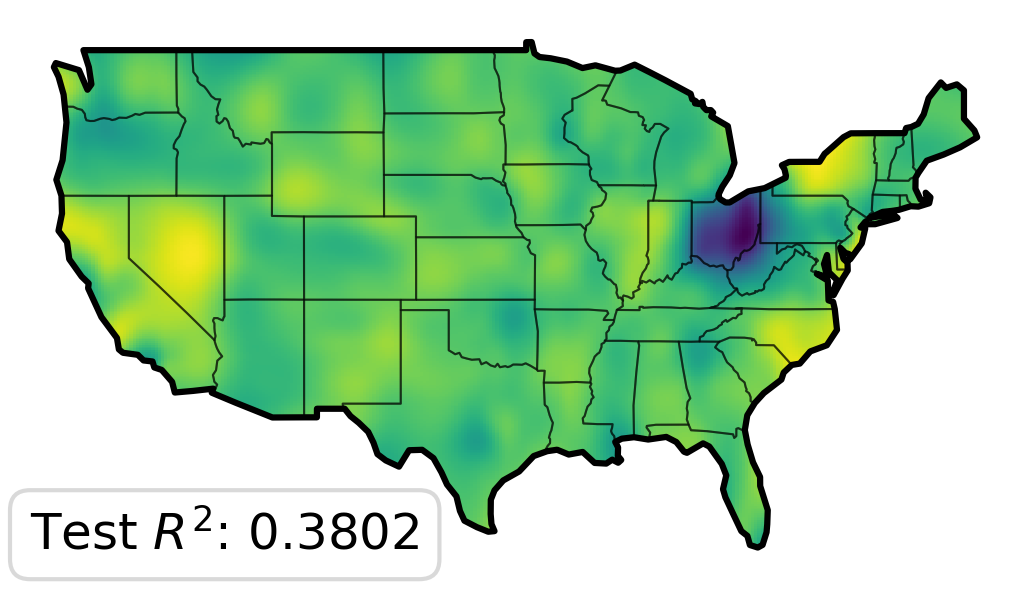}
  \includegraphics[width=0.23\textwidth]{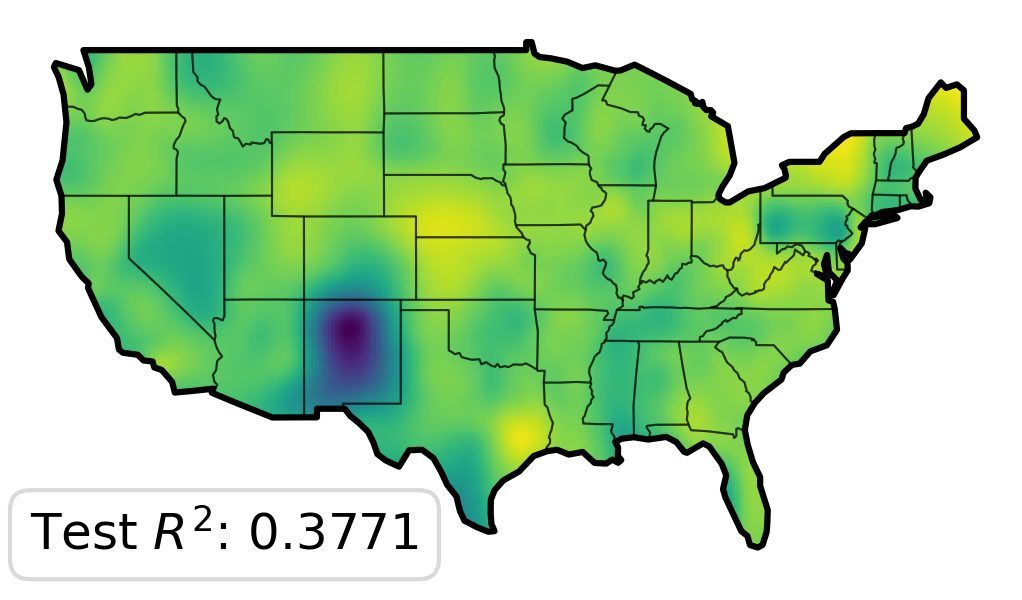}
  \includegraphics[width=0.23\textwidth]{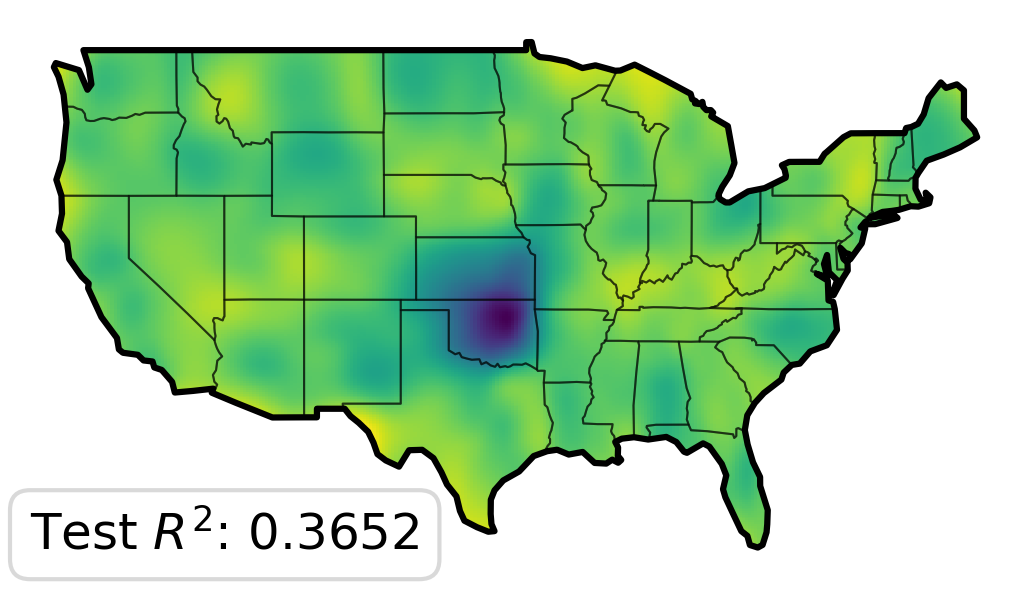}
  \includegraphics[width=0.23\textwidth]{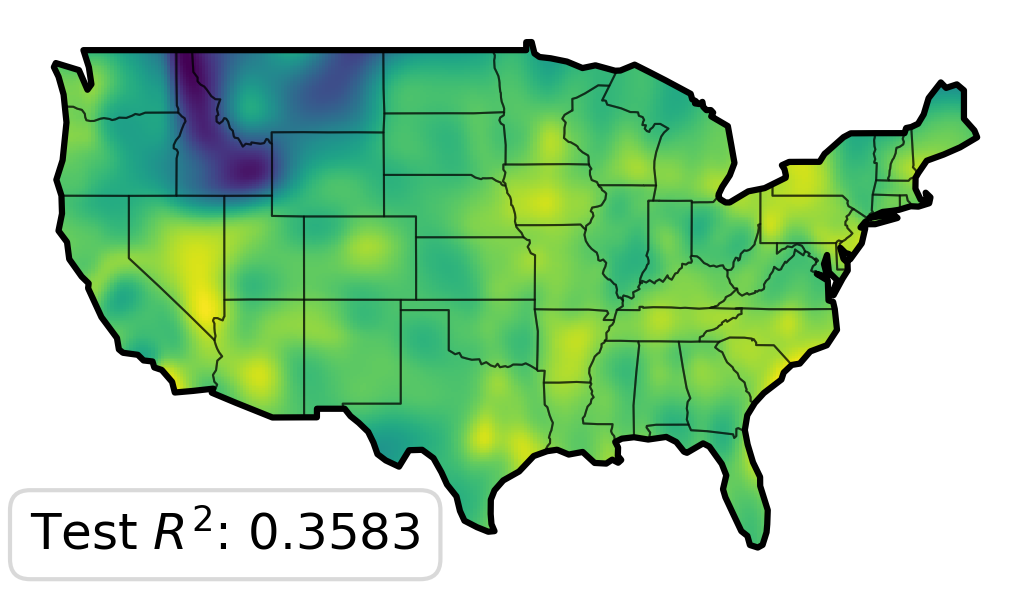}
  \includegraphics[width=0.23\textwidth]{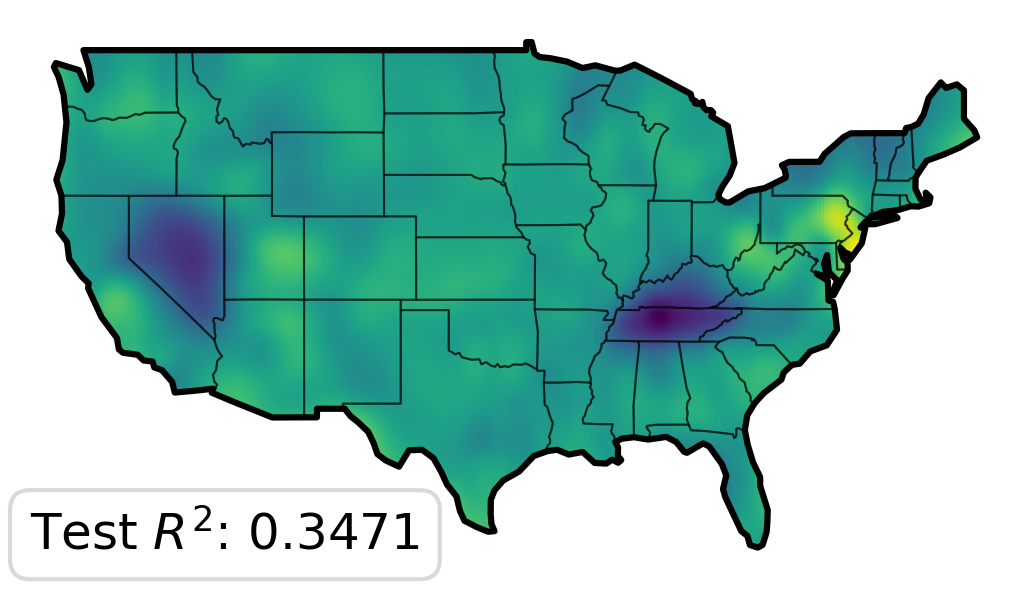}
  \includegraphics[width=0.23\textwidth]{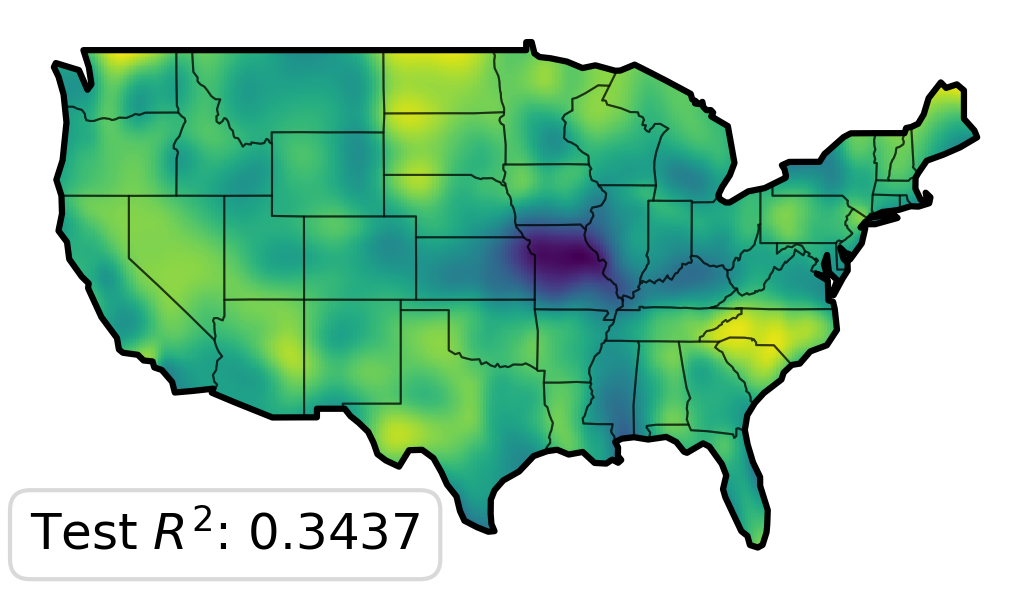}
  \includegraphics[width=0.23\textwidth]{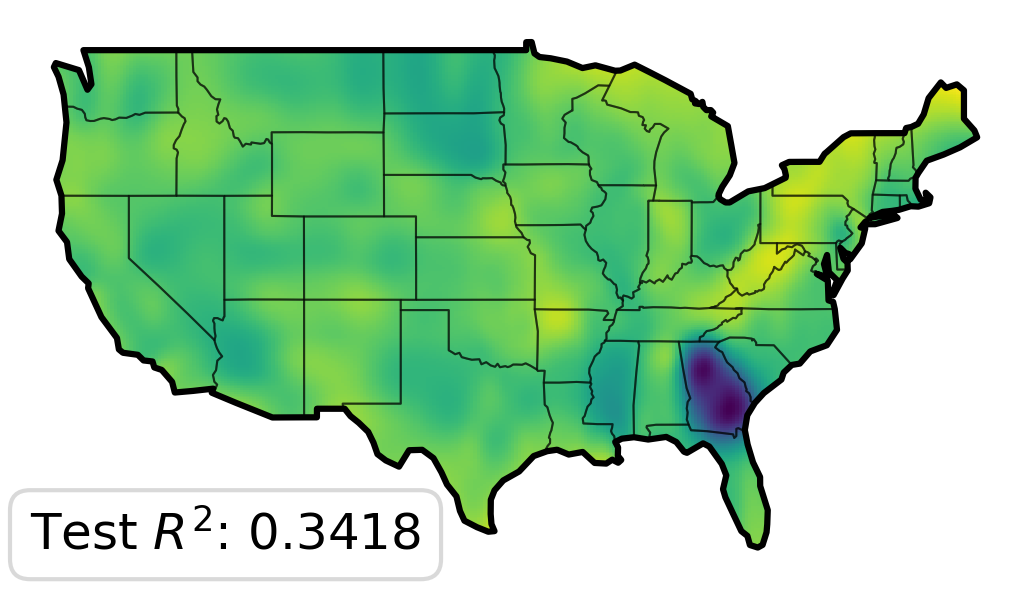}
  \includegraphics[width=0.23\textwidth]{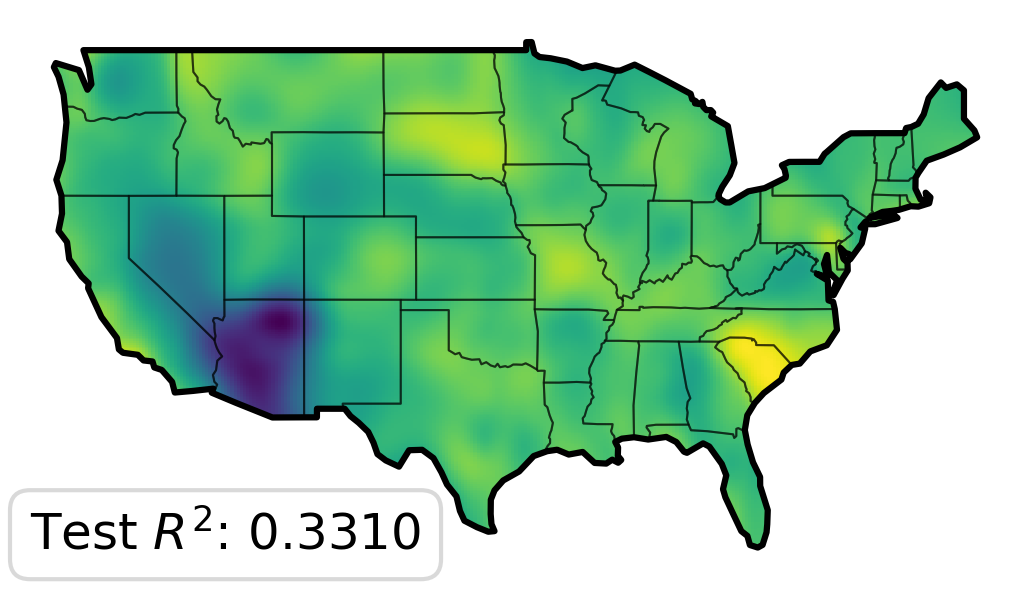}
  \includegraphics[width=0.23\textwidth]{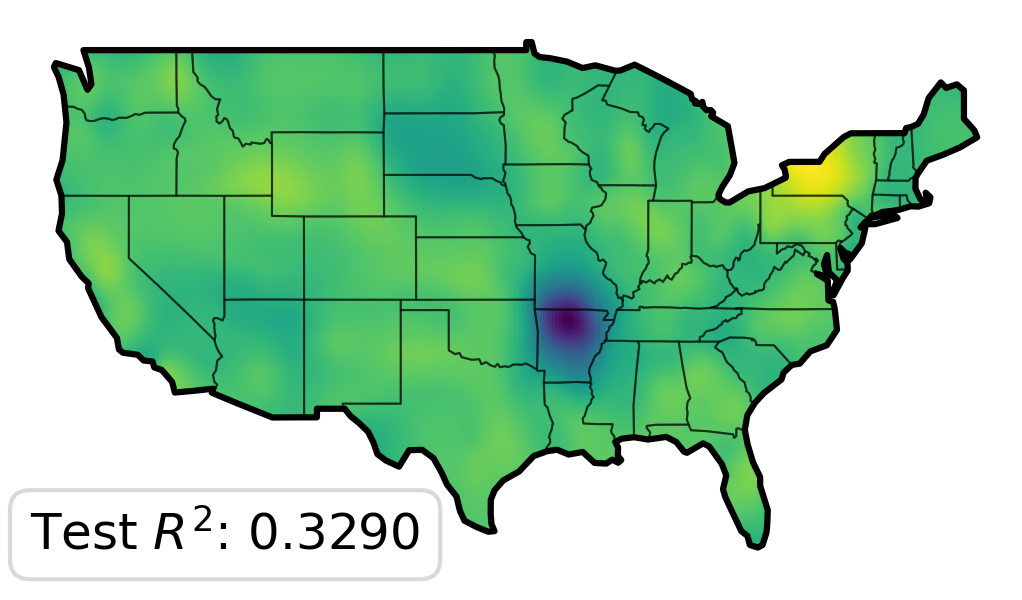}
  \includegraphics[width=0.23\textwidth]{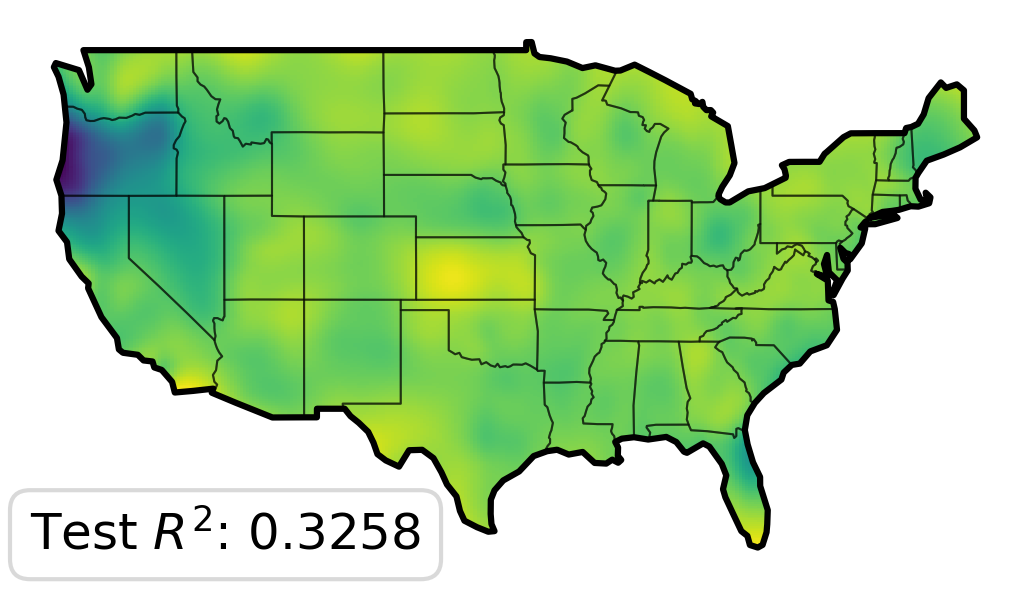}
  \includegraphics[width=0.23\textwidth]{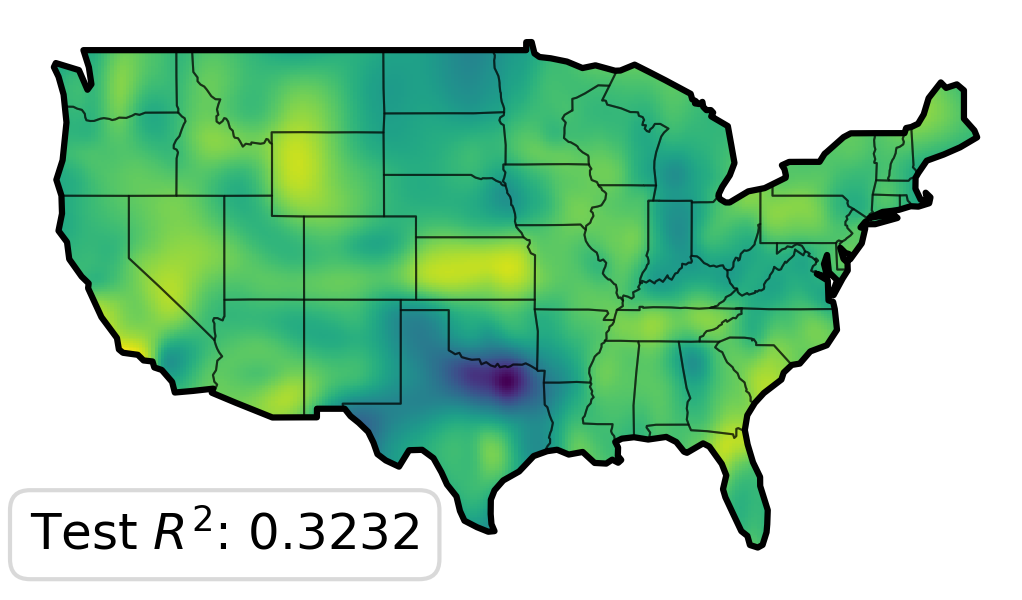}
  \includegraphics[width=0.23\textwidth]{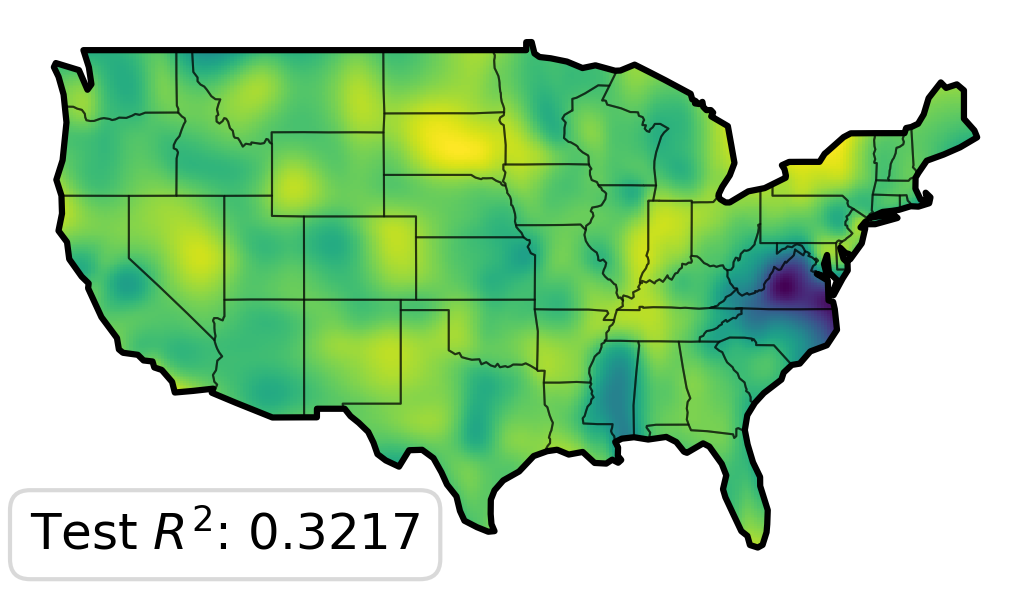}
  \includegraphics[width=0.23\textwidth]{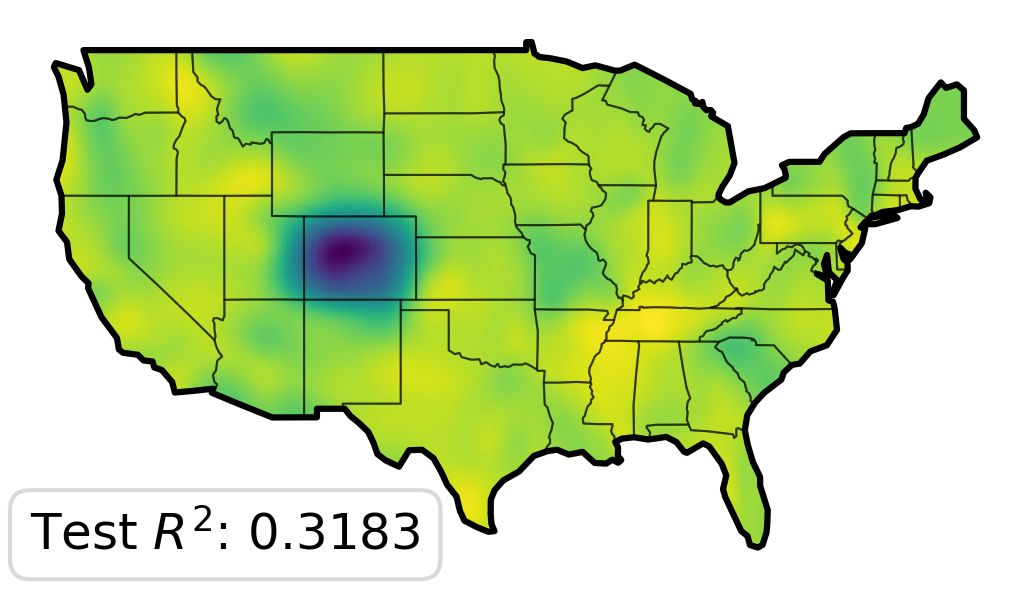}
  \includegraphics[width=0.23\textwidth]{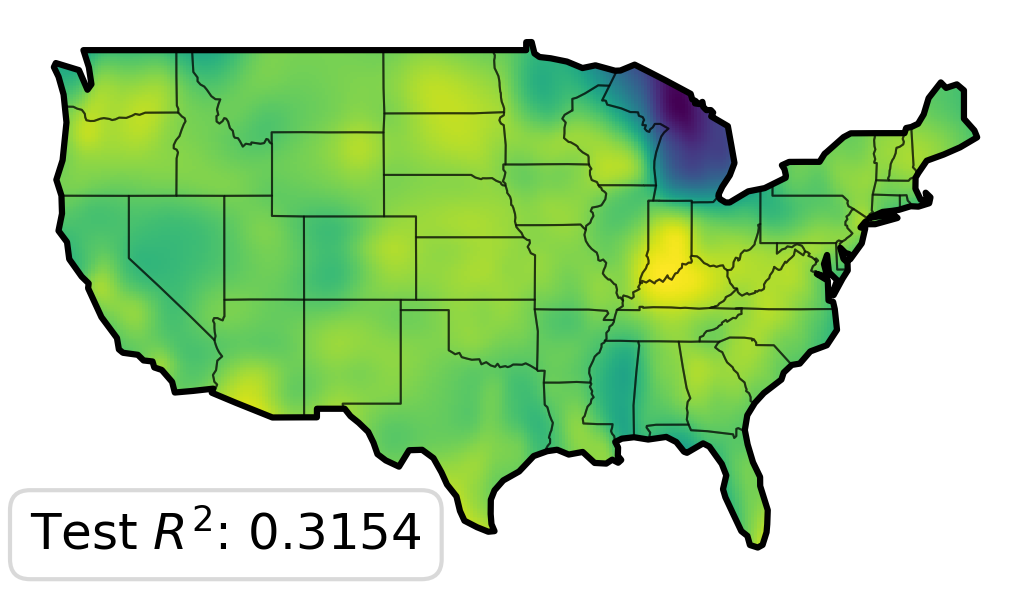}
  \includegraphics[width=0.23\textwidth]{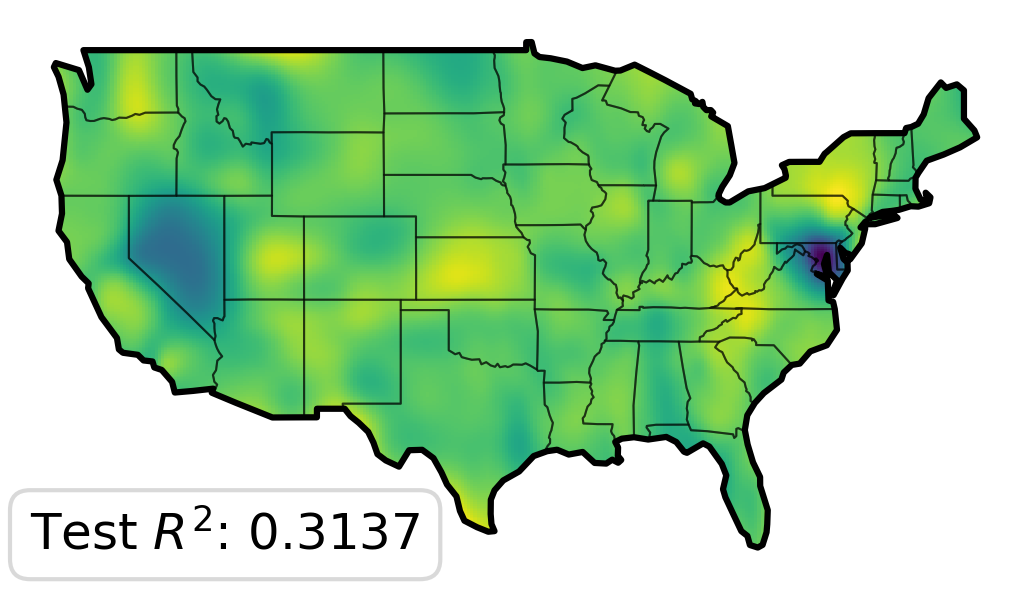}
  \includegraphics[width=0.23\textwidth]{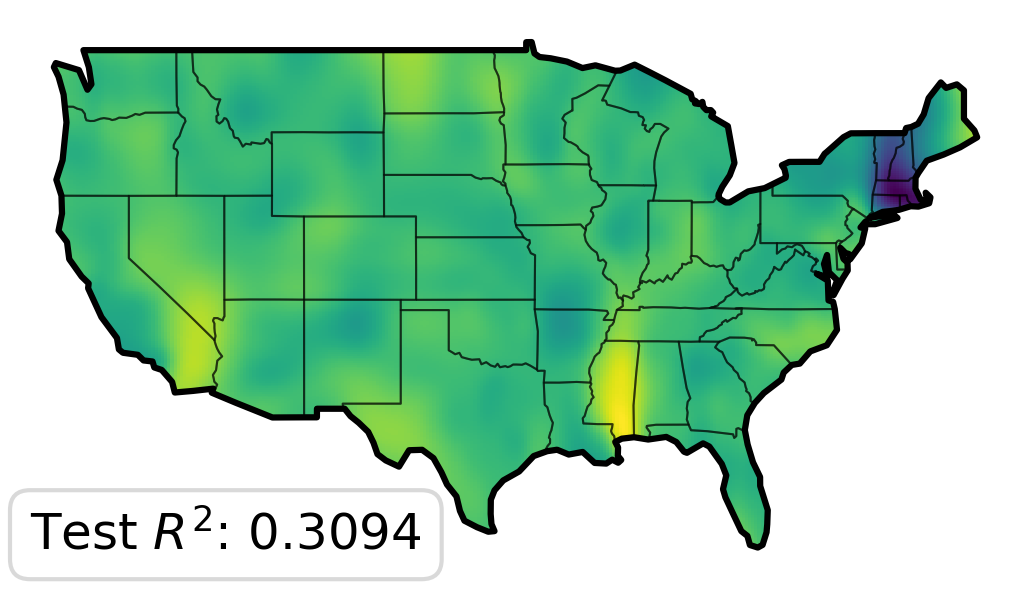}
  \includegraphics[width=0.23\textwidth]{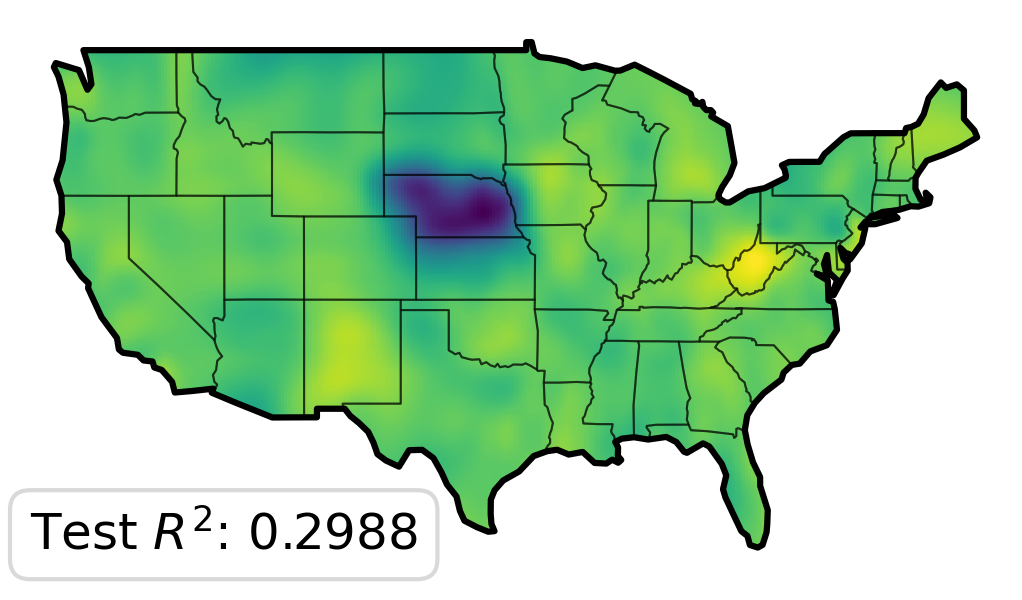}
  \includegraphics[width=0.23\textwidth]{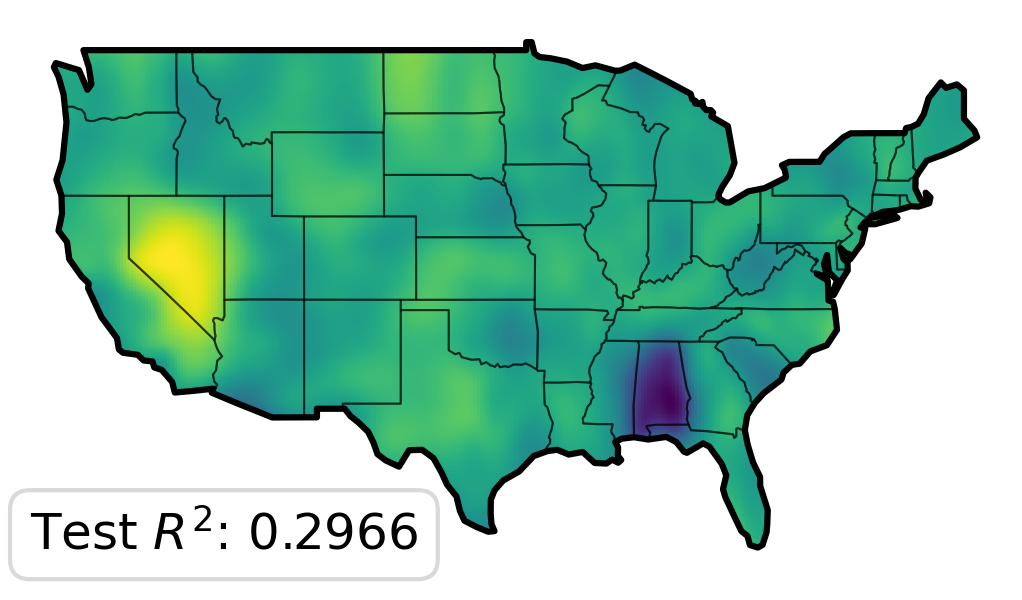}
  \includegraphics[width=0.23\textwidth]{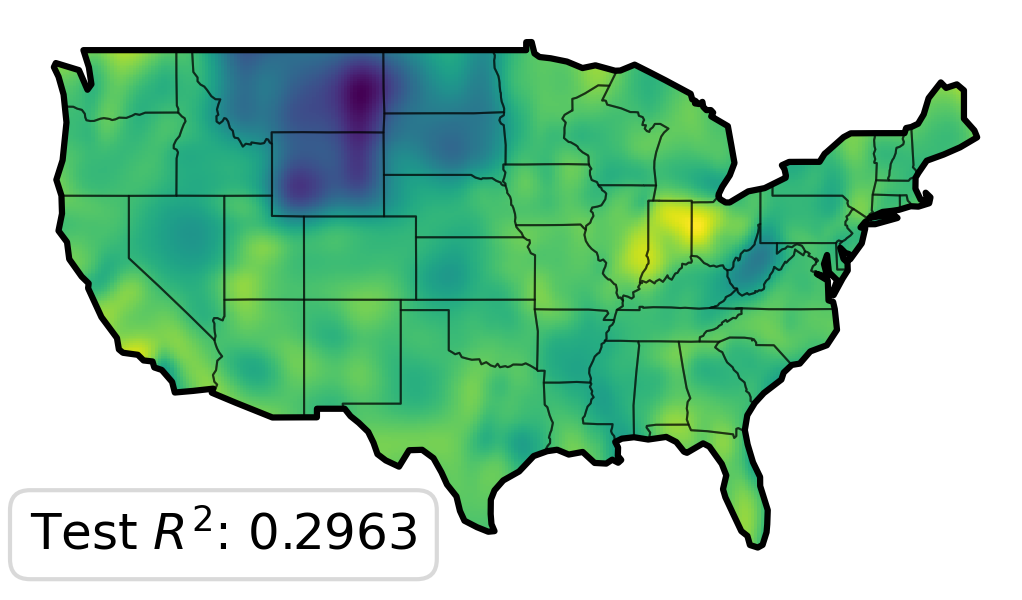}
  \includegraphics[width=0.23\textwidth]{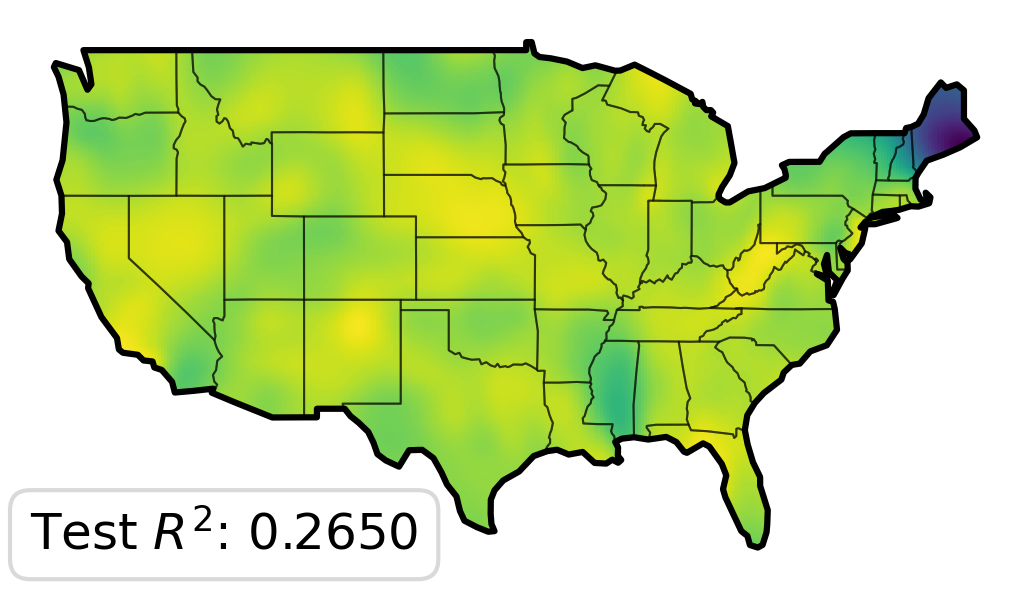}
  \includegraphics[width=0.23\textwidth]{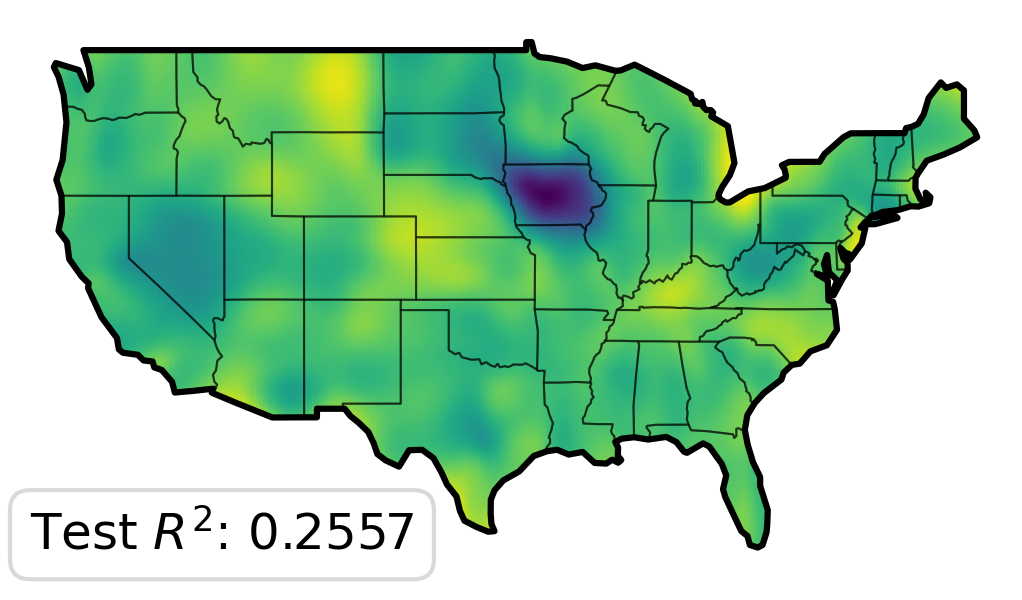}

  \caption{The top 5 time features, and top 32 space features from layer 16 of Llama 2-7b after applying a Varimax rotation to make them approximately sparse. The rotation makes the features interpretable. In particular, many of the space features localize on particular U.S. states, and the time features separate the decades from the 1950s to the 2010s.}
  
  \label{fig:varimax}
\end{figure}

\begin{figure}[p]
  
  \centering

  \includegraphics[width=\textwidth]{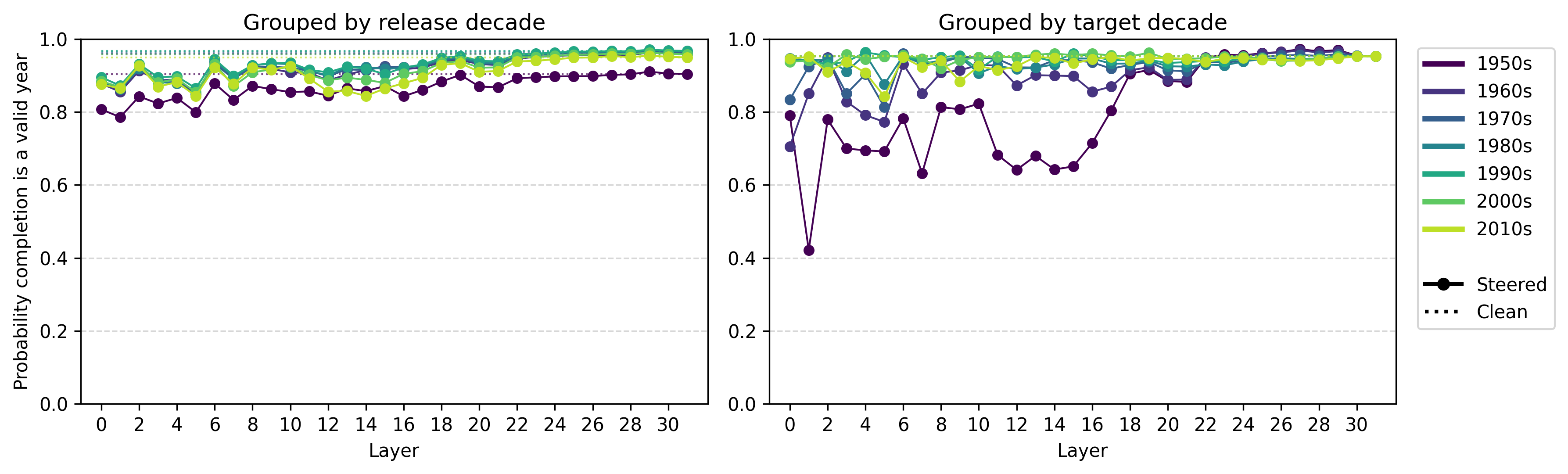}

  \caption{The mean probability that the model completes the prompt with a valid year in the steering experiment, grouped by release decade (left) and target decade (right). The dashed line shows the mean probability for the clean runs. The interventions have very little effect on the model's ability to meaningfully complete the prompt, with the exception of steering to years in the 1950s.}
  \label{fig:degradation}
\end{figure}

\begin{figure}[p]
  \centering
  
  \includegraphics[width=\textwidth]{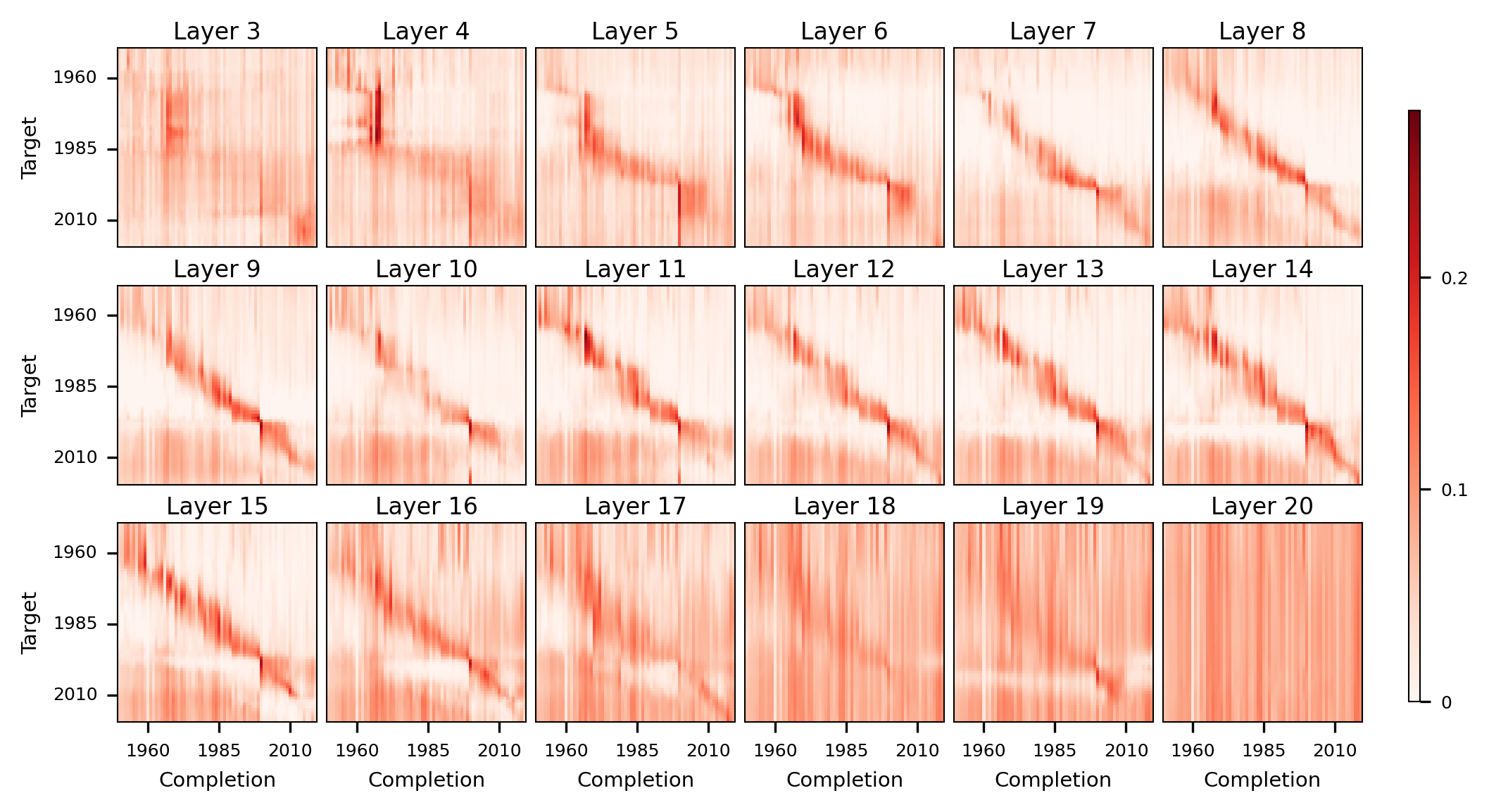}
  
  \caption{Colour intensity indicates the standard deviation of the probability of a completion given the steering target in the steering experiment.}
  \label{fig:steering_std}
\end{figure}

\FloatBarrier

\section{An efficient algorithm to fit the regularization parameters}
\label{sec:selecting_regularization_parameters}

In this section, we discuss an optimization strategy which allows us to efficiently optimize the Manifold Probe objective in \eqref{eq:probe_f} while also selecting the regularization parameters $\lambda_w$ and $\lambda_f$ using a closed-form criterion such as GCV or REML which apply to linear predictors.

Instead of directly employing the closed-form solution in Proposition~\ref{prop:closed_form_solution}, we propose the Alternating Least Squares procedure detailed in Algorithm~\ref{alg:als}. 
Here, we have used the notation $\|\alpha\|_{\Sigma} = \sqrt{\alpha^\top \Sigma \alpha}$ and write $\alpha \perp_{\Sigma} \beta$ to mean $\alpha^\top \Sigma \beta = 0$.

\IncMargin{.5cm}

\begin{algorithm}[htbp]
\caption{Alternating least squares optimization of \eqref{eq:probe_f}.}
\label{alg:als}

\SetAlFnt{\fontfamily{cmss}\selectfont}

\SetKwSty{textcmrbold}

\SetArgSty{textcmr}

\SetKwRepeat{Repeat}{repeat:\vspace{0.2em}}{until}
\SetKwFor{For}{for\vspace{0.2em}}{}

\vspace{0.2em}
\hspace{-.575cm} \textcmrbold{Input:} \textcmr{inital parameters $\beta^{(0)}_1,\ldots,\beta^{(0)}_d$}\;
\vspace{0.3em}

\For{$k = 1, \ldots, d$\textcmrbold{:}}{
\For{$t=1,2,\ldots$\textcmrbold{:}}{
    \textcmr{\textbf{$w$-update}}:\\
    \vspace{0.3em}
    $\quad w^{(t+1)} \longleftarrow \;\;\;\underset{w}{\operatorname{argmin}}\;\;\; \|y - X w\|^2_2 + \tilde\lambda_w \|w\|_2^2$, \hspace{2.15em} $y = H \beta^{(t)}$\;
      \vspace{0.4em}
      \textcmrbold{$\beta$-update}:\vspace{0.3em}\\
      $\quad\beta^{(t+\sfrac{1}{2})} \longleftarrow \underset{\beta \perp_{\Sigma} \hat \beta_{k-1}, \ldots, \hat \beta_1}{\operatorname{argmin}} \|y - H\beta\|^2_2 + \tilde\lambda_f \beta^\top S \beta$, \hspace{2em} $y = X w^{(t+1)}$\;
      \vspace{0.3em}
      $\quad\beta^{(t+1)} \longleftarrow \beta^{(t+\sfrac{1}{2})} / \|\beta^{(t+\sfrac{1}{2})}\|_\Sigma$\;
    \vspace{0.3em}
}
}

\vspace{0.3em}

\end{algorithm}

\DecMargin{.5cm}

While it may seem needlessly inefficient to optimize \eqref{eq:probe_f} using Algorithm~\ref{alg:als} rather than the closed-form solution in Proposition~\ref{prop:closed_form_solution}, when we formulate the power-iteration procedure used to solve the generalized eigenvalue problem, we see that this is exactly equivalent to the alternating least squares procedure. Power-iteration is known to converge to the global solution under mild conditions on the initial value, and therefore we can guarantee that Algorithm~\ref{alg:als} converges to the global solution under the same conditions.

\begin{lemma}
  \label{lem:als_convergence}
  Suppose that $\hat\nu_{k} > \hat \nu_{k+1}$ and $w_k^{(0)} \not \perp \hat w_k$ for all $k=1,\ldots,d$, then for some $\tilde \lambda_w, \tilde \lambda_f$, Algorithm~\ref{alg:als} converges to the global minimizer of \eqref{eq:probe_f}. I.e.
  \[
    \lim_{t\to\infty} f^{(t)}_k = \hat f_k, \qquad \text{ and } \qquad \lim_{t\to\infty} w^{(t)}_k = \hat w_k
  \]
\end{lemma}

The distinct eigenvalue condition is not strictly necessary and can be relaxed to simply $\hat \nu_d > \hat \nu_{d+1}$, allowing repeated eigenvalues. The stricter condition is stated for simplicity. We provide a proof based on the power-iteration argument in Section~\ref{sec:proof_of_als_convergence}.

By framing the optimization in this way, we can apply closed-form criteria designed for linear predictors such as GCV or REML to select the regularization parameters $\tilde \lambda_w$ and $\tilde \lambda_f$ at each iteration of the alternating least squares procedure. 

Viewed this way, we are also not restricted to quadratic penalties, and can use non-quadratic penalties such as the $\ell_1$ or elastic-net-type penalties, provided we have an efficient off-the-shelf regression solver which accomodates it.

\subsection{Efficient parametrization of the ridge regression problems}
\label{sec:efficient_optimization_of_als}

To efficiently perform the required computations in Algorithm~\ref{alg:als}, we perform some reparametrizations and matrix decompositions which allow us reduce each iteration to simple matrix multiplcations of size $p \times p$, removing the time-dependence on the number of samples and avoiding performing any matrix inversions.

To do this, we first reparametrize the $\beta$-problem to enforce the linear orthogonality constraints, and then reparametrize it again so that $S$ becomes the identity matrix. If $S$ is rank-deficient, we simply set its zero eigenvalues to some small positive constants to make it positive-definite to allow the reparametrization. We are then left with a standard ridge regression problem. We note that from here on we parametrize and solve the $w$-problem in exactly the same way, so we won't discuss it separately.

We next compute the singular value decomposition of $H$ as $H = U D V^\top$, where all diagonal entries of $D$ are positive, and reparametrize the problem again to make $H = UD$. This ensures that $H$ has full-column rank and avoids some unnecessary matrix multiplications down the line. It is then straightforward to show that
\begin{equation}
  \label{eq:p_dim_ridge}
  \|y - H \beta\|_2^2 + \lambda \|\beta\|_2^2 = \|\mathbbm{y} - D \beta\|_2^2 + \lambda \|\beta\|_2^2 + r
\end{equation}
where $\mathbbm{y} = U^\top y$ and $r = \|y\|_2^2 - \|\mathbbm{y}\|_2^2$ is a constant which does not depend on $\beta$. Note that $\mathbbm{y} = U^\top Xw$ so as long as $U^\top X$ is pre-computed, this multiplication does not depend on the number of samples $n$. The solution to the ridge regression problem \eqref{eq:p_dim_ridge} is
\[
  \hat \beta = (D^2 + \lambda I)^{-1} D \mathbbm{y}
\]
which given $\mathbbm{y}$ can be computed in $O(p)$ time for any $\lambda$. The GCV or REML criterion and their gradients and Hessians can also be computed efficiently using this reparametrization, allowing us to very efficiently select the regularization parameters at each iteration using Newton's method.

\subsection{Proof of Lemma~\ref{lem:als_convergence}}
\label{sec:proof_of_als_convergence}

We begin by showing the convergence of $\beta^{(t)}$ for $k=1$. Once this is established, the convergence of $w^{(t)}$ is trivial. We'll consider the case that $k=1$ and note that the subsequent cases follow by a deflation argument.

We recall that the ridge updates have closed forms
\[
  w^{(t+1)} = (X^\top X + \tilde \lambda_w I)^{-1}X^\top H\beta^{(t)}
\]
and
\[
  \beta^{(t+\sfrac{1}{2})} = (H^\top H + \tilde \lambda_f S)^{-1} H^\top X w^{(t+1)} = (H^\top H + \tilde \lambda_f S)^{-1} H^\top X (X^\top X + \tilde \lambda_w I)^{-1} X^\top H \beta^{(t)}.
\]
We define the matrix $T = LH^\top A H$ where $L = (H^\top H + \tilde \lambda_f S)^{-1}$ and $A = X (X^\top X + \tilde \lambda_w I)^{-1} X^\top$, so that
\[
  \beta^{(t+\sfrac{1}{2})} = T \beta^{(t)}.
\]
A full $\beta$-update is then given by
\[
  \beta^{(t+1)} = \frac{T\beta^{(t)}}{\|T\beta^{(t)}\|_{\Sigma}}.
\]
where $\|a\|_{\Sigma} = \sqrt{a^\top \Sigma a}$ with $\Sigma = H^\top H/n$.
This shows that the sequence is a power-iteration for the matrix $T$, and therefore as long as its largest eigenvalue is unique, and the initial value $\beta^{(0)}$ is not orthogonal to the corresponding eigenvector, it converges to the leading eigenvector of $T$ by a standard argument\footnote{see, for example, {\tt https://en.wikipedia.org/wiki/Power\_iteration}.}. It remains to show that the leading eigenvector of $T$ is the same as the eigenvector $\hat \beta$ with the smallest eigenvalue $\hat \nu$ of the generalized eigenvalue problem 
\[
  M \hat \beta = \hat \nu \Sigma \hat\beta
\]
Plugging in $M$ and $\Sigma$ and rearranging we obtain
\[
  H^\top A H \hat \beta = \left[ (1 - \hat \nu/n) H^\top H + \tilde \lambda_f S \right] \hat \beta
\]
and setting $\tilde \lambda_f = \lambda_f / (1 - \hat \nu/n)$, we have
\[
  H^\top A H \hat \beta = (1 - \hat \nu/n)\left( H^\top H + \tilde \lambda_f S \right) \hat \beta =: (1 - \hat \nu/n) L^{-1} \hat \beta.
\]
Multiplying both sides on the left by $L$, we have
\[
  T \hat \beta = (1 - \hat \nu / n)\hat \beta
\]
which shows that $\hat \beta$ is the leading eigenvector of $T$, which completes the proof.

\section{Proof of Lemma~\ref{lem:population_regression}}
\label{sec:proof_of_population_regression}

\subsection{Proof of \eqref{eq:f_objective}}

Let $f_1^\star, \ldots, f_d^\star$ be any set of mean-zero, orthonormal features which satisfy \eqref{eq:decomposition}, and let $\*F = \Span\cu{f_1^\star, \ldots, f_d^\star}$. We will show that $\Span\{f_1,\ldots, f_d\} = \*F$.
We begin with the case $k=1$. By the law of iterated expectation, \eqref{eq:f_objective} can be written as
    \[
        \E \sq*{\br*{f(z) - w^\top x - b}^2} = \E \sq*{ \E \sq*{\br*{f(z) - w^\top x - b}^2 \mid z}} 
    \]
    therefore
    \[
        f_1(z) = \mathbb E\sq*{w^\top x + b \mid z} = w^\top\mathbb E\sq*{x \mid z} + b.
    \]
    Now
    \[
        E\sq*{x \mid z} = \phi(z) + \E\sq*{\eta} = \phi(z) + \bar \eta
    \]
    and so
    \[
        f_1(z) = w^\top \sq*{\phi(z) + \bar \eta} + b = w^\top \br*{f_1^\star(z) u_1 + \cdots f_d^\star(z) u_d} = (w^\top u_1) f_1^\star(z) + \cdots + (w^\top u_d) f_d^\star(z) + w^\top \bar \eta + b.
    \]
    Now, $f_1$ is constrained so that $\E[f_1] = 0$, and given that $\E[f_1^\star] = \cdots = \E[f_d^\star] = 0$, this implies that $w^\top \bar \eta + b = 0$. Therefore, $f_1$ is a linear combination of $f_1^\star,\ldots, f_d^\star$. Since $\E(f^2) = 1$, at least one coefficient must be non-zero and therefore $f_1 \in \*F$.

    Next, we suppose that $f_1,\ldots, f_{k-1} \in \*F$ and $f_1 \perp \cdots \perp f_{k-1}$ for some $k \in \cu*{1,\ldots, d}$. We will show that $f_k \in \*F$.

    Minimizing \eqref{eq:f_objective} subject to the constraint $f_k \perp f_{k-1}, \ldots, f_1$ is equivalent to minimizing \eqref{eq:f_objective} replacing $x(z)$ with the deflation
    \[
        x^{(k)}(z) = x(z) - \br*{\pi_1 f_1(z) + \cdots + \pi_{k-1}f_{k-1}(z)}, \qquad \pi_i = \E\sq*{x f_i}.
    \]
    As before, this is minimized by
    \[
        f_k(z) = w^\top \E\sq*{x^{(k)} \mid z} + b = w^\top \sq*{\phi(z) + \bar \eta} + b - \br*{\pi_1 f_1(z) + \cdots + \pi_{k-1}f_{k-1}(z)}
    \]
    \[
        = (w^\top u_1)f_1^\star(z) + \cdots + (w^\top u_d) f_d^\star(z) - \br*{\pi_1 f_1(z) + \cdots + \pi_{k-1}f_{k-1}(z)}.
    \]
    Since this is a linear combination of functions in $\*F$, and $f_k$ is constrained to be non-trivial, this implies that $f_k \in \*F$. Therefore, by induction, $f_1,\ldots, f_d \in \*F$.

    Now the functions $f_1,\ldots, f_d$ are constrained to be orthogonal, and therefore the span $\*F$, so
    \[
        \Span\cu{f_1,\ldots, f_d} = \Span\cu{f_1^\star, \ldots, f_d^\star}.
    \]

    \subsection{Proof of \eqref{eq:u_objective}}

  To find the optimal $(u_k, c_k)$ that minimizes the population least-squares error
  \[
    \mathbb E\left[\left\|x - u f_k(z) - c\right\|^2\right]
  \]
  we start by taking the gradient with respect to $c$ and setting it equal to zero to obtain
  \[
    -2 \mathbb E\left[x - u f_k(z) - c\right] = 0
  \]
  which gives that
  \[
    c = \bar x - u\E[f_k] = \bar x
  \]
  where $\bar x = \mathbb E[x]$ and we have used the fact that $\E[f_k] = 0$. Substituting this back into \eqref{eq:u_objective}, taking the gradient with respect to $u$ and setting it equal to zero gives us 
  \[
    -2 \mathbb E\left[f_k(z)(x - \bar x - u f_k(z))\right] = 0
  \]
  which implies that
  \[
    \E\sq*{f_k(z)(x - \bar x)} = u \E[f_k^2].
  \]
  Since $\E[f_k^2] = 1$, this gives us that
  \[
    u = \E\sq*{f_k(z)(x - \bar x)}.
  \]
  Now $\bar x = \bar \eta$, and so
  \[
    u = \E\sq*{f_k(z)(\phi(z) + \eta(\xi) - \bar \eta)} = \E\sq*{f_k(z) \phi(z)} + \E\sq*{f_k(z)\eta(\xi)} - \E\sq*{f_k(z) \bar \eta} = \E\sq*{f_k(z) \phi(z)}
  \]
  where the final two terms are zero by the assumption of independence between $z$ and $\xi$, and the fact that $\E[f_k] = 0$. Now, substituting in the decomposition of $\phi(z)$ gives us
  \[
    u = \E\sq{f_k(z)\br{u_1f_1(z) + \cdots + u_d f_d(z)}} = u_1 \E[f_k f_1] + \cdots + u_d \E[f_k f_d]
  \]
  Now by assumption $\E[f_k f_j] = 0$ for all $j \neq k$, and $\E[f_k^2] = 1$, so we have that
  \[
    u = u_k,
  \]
  as required.

\section{Proof of Proposition~\ref{prop:closed_form_solution}}
\label{sec:proof_of_closed_form_solution}

We begin by observing that since $f\in\*H$, it can be written as $f(z) = \beta^\top h(z)$ and the constraint $\sum_{i=1}^n f(z_i) = 0$ implies that $\beta^\top \bar h = 0$. Therefore, we can write evaluations of $f$ in terms on the centered model matrix $H$:
\[
  f(z_i) = \beta^\top h(z_i) = \beta^\top(h(z_i) - \bar h) = (H\beta)_i.
\]

Now, to minimize \eqref{eq:probe_f} with respect to $b$, we take the gradient with respect to $b$ and set it equal to zero to obtain
\[
  -2\sum_{i=1}^n \left(f(z_i) - w^\top x_i - b\right) = 0.
\]
Since $\sum_{i=1} f(z_i) = 0$, we obtain that $b = -w^\top \bar x$. Therefore we can write
\[
  f(z_i) - w^\top x_i - b = f(z_i) - w^\top x_i - (-w^\top \bar x) = f(z_i) - w^\top (x_i - \bar x) = (H\beta)_i - (Xw)_i.
\]
Recalling that $J(f) = \beta^\top S \beta$, this means we can write \eqref{eq:probe_f} in matrix form as minimizing
\begin{equation}
  \label{eq:matrix_objective}
  \|H\beta - Xw\|_2^2 + \lambda_w w^\top w + \lambda_f \beta^\top S \beta
\end{equation}
subject to $\beta^\top \Sigma \beta = 1$ and $\beta^\top \Sigma \hat \beta_j = 0$ for all $j < k$. 

For fixed $\beta$, the optimal $w$ is given by the ridge regression solution
\[
  w = (X^\top X + \lambda_w I)^{-1} X^\top H \beta.
\]

With $A = X^\top(X^\top X + \lambda_w I) X^\top$ as in the proposition, we have
\[
        \begin{aligned}
            \norm*{H \beta - X w}^2_2 &= \norm*{H\beta - X (X^\top X + \lambda_w I)^{-1} X^\top H \beta}_2^2 \\
            &= \norm*{(I - A) H \beta}_2^2 \\
            &= \beta^\top H^\top (I - A)^2 H \beta \\
            &= \beta^\top H^\top(I - 2A + A^2)H \beta.
        \end{aligned}
    \]

Now, we define $K = (X^\top X + \lambda_w I)^{-1}$, and noting that $X^\top X = K^{-1} - \lambda_w I$, we observe that
    \[
        \begin{aligned}
            A^2 &= XK X^\top X K X^\top \\
            &= XK(K^{-1} - \lambda I) K X^\top \\
            &= X K K^{-1} K X^\top - \lambda_w X K^2 X^\top \\
            &= XK X^\top - \lambda_w X K^2 X^\top \\
            &= A - \lambda_w X K^2 X^\top 
        \end{aligned}
    \]
    and
    \[
        \|w\|^2 = \|K X^\top H \beta\|_2^2 = \beta^\top H^\top X K^2 X^\top H \beta.
    \]

Therefore
\[
  \beta^\top H^\top A^2 H \beta = \beta^\top H^\top A H \beta - \lambda_w \beta^\top H XK^2X H^\top \beta = \beta^\top H^\top A H \beta - \lambda_w \|w\|^2,
\]
and it follows that
\[
  \|H\beta - Xw\|_2^2 + \lambda_w \|w\|_2^2 = \beta^\top H^\top (I - A) H \beta.
\]

Therefore, the objective function \eqref{eq:matrix_objective} can be written as
\[
  \|H\beta - Xw\|_2^2 + \lambda_w \|w\|_2^2 + \lambda_f \beta^\top S \beta = \beta^\top \left[ H^\top (I - A) H + \lambda_f S \right] \beta =: \beta^\top M \beta.
\]
By the Rayleigh-Ritz theorem, minimizer of this with respect to $\beta$, subject to $\beta^\top \Sigma \beta = 1$ is given by the generalized eigenvector of \eqref{eq:eval_eq} with the smallest eigenvalue. Continuing sequentially, the generalized eigenvector with the $k$th smallest eigenvalue gives the solution to \eqref{eq:matrix_objective} subject to $\beta^\top \Sigma \hat \beta_j = 0$ for all $j < k$. This completes the proof.

\end{document}